\relax

\documentclass[preprint]{elsarticle}

\usepackage{times}  
\usepackage{helvet}  
\usepackage{courier}  
\usepackage[hyphens]{url}  
\usepackage{graphicx} 
\usepackage[numbers]{natbib}  
\usepackage{caption} 
\def\BibTeX{{\rm B\kern-.05em{\sc i\kern-.025em b}\kern-.08em
    T\kern-.1667em\lower.7ex\hbox{E}\kern-.125emX}}
\usepackage{subcaption}
\usepackage{colortbl}
\usepackage{graphicx}
\usepackage{stfloats}
\usepackage{booktabs}
\usepackage{flushend}
\usepackage{multirow}
\usepackage{amssymb}
\usepackage{wrapfig}
\usepackage{makecell}
\usepackage{bm}
\usepackage{amsmath}

\usepackage{ragged2e}
\usepackage{float}
\usepackage{xcolor}
\usepackage[normalem]{ulem} 
\newcommand\hl{\bgroup\markoverwith{\textcolor{pink}{\rule[-.5ex]{2pt}{2.5ex}}}\ULon}

\makeatletter
\renewcommand{\maketag@@@}[1]{\hbox{\m@th\normalsize\normalfont#1}}%

\makeatother
\begin{document}
\begin{frontmatter}%

\title{A-SFS: Semi-supervised Feature Selection based on Multi-task Self-supervision}
\author[inst1]{Zhifeng Qiu}

\author[inst1]{Wanxin Zeng}
\author[inst1]{Dahua Liao}
\author[inst2]{Ning Gui*}

\address[inst1]{School of Automation, Central South University,
            Changsha,
            {410083}, 
            {Hunan},
            {China}}
            
\address[inst2]{School of Computer Science, Central South University,
            {Changsha},
            {410083}, 
            {Hunan},
            {China}}



\begin{abstract}


%

Feature selection is an important process in machine learning. It builds an interpretable and robust model by selecting the features that contribute the most to the prediction target. However, most mature feature selection algorithms, including supervised and semi-supervised, fail to fully exploit the complex potential structure between features. We believe that these structures are very important for the feature selection process, especially when labels are lacking and data is noisy. 

To this end, we innovatively introduces a deep learning-based self-supervised mechanism into feature selection problems, namely batch-Attention-based Self-supervision Feature Selection(A-SFS). Firstly, a multi-task self-supervised autoencoder is designed to uncover the hidden structural among features with the support of two pretext tasks. Guided by the integrated information from the multi-self-supervised learning model, a batch-attention mechanism is designed to generate feature weights according to batch-based feature selection patterns to alleviate the impacts introduced from a handful of noisy data. This method is compared to 14 major strong benchmarks, including LightGBM and XGBoost. Experimental results show that A-SFS achieves the highest accuracy in most datasets. Furthermore, this design significantly reduces the reliance on labels, with only 1/10 labeled data are needed to achieve the same performance as those state of art baselines. Results show that A-SFS is also most robust to the noisy and missing data.

\end{abstract}




\begin{keyword}
Feature selection, attention mechanism, self-supervised , autoencoder.
\end{keyword}

\end{frontmatter}

\section{Introduction}
\label{sec:Introduction}

With the prompt advancement of the Internet of Things(IoT) and Industrial Automation Systems(IAS), enterprises and industries collect and accumulate data with unparalleled speed and volume.
In order to extract valuable information from the large quantity of otherwise meaningless data,  
one crucial machine-learning process is feature selection (FS)~\citep{yin2014review}, which focuses on removing irrelevant, redundant, and noisy features concerning the supervision target. Simplified feature subsets can usually retain most of the vital physical meaning of the raw features, making the model more readable and interpretable.

Existing feature selection methods for conventional data are based on a strong assumption that features are independent of each other (flat) while ignoring the inherent feature structures\citep{li2017feature}. However, existing data normally have latent structures, spatial or temporal smoothness, groups\citep{jacob2009group}, trees and graphs\citep{zhao2015graph} or even with heterogeneous structures\citep{zhou2021feature}. For complex datasets with those latent  structures, many feature selection solutions are facing the following challenges: 

\noindent\textbf{Complex latent structure vs. Predefined evaluation criteria:}
Current methods typically propose evaluation criteria that only focus on a specific type of merit, e.g., 
 information gain~\citep{uuguz2011two}, joint mutual information~\citep{bennasar2015feature}, double sparse learning~\citep{li2020survey}. In reality, the structure between different features could be very complex and diverse, with possible combinations of different relations~\cite{gui2017Feature}. Such predefined criteria can hardly effectively describe the complex relationships among features.

\noindent\textbf{High data diversity vs. Solutions with handcrafted domain-specific model:}
In order to achieve the optimal FS performance, many domain-specific feature selection methods are proposed by integrating domain-specific knowledge related to the feature structures, most of which are supported with handcrafted domain-specific models, e.g.,\citep{alweshah2021coronavirus} in medical and \citep{li2021wind} in energy. Those models are highly reliant on the human-experts, hard to construct, and applicable in certain domains. Thus, they are limited in transferability and reusability. Due to the high data diversity, a good FS solution should be able to uncover the latent structures directly from the data themselves.


\noindent\textbf{Label scarcity vs. High label dependence:} Existing FS methods normally highly rely on labels for feature selection. In the case of label scarcity, they are likely to suffer significant performance deterioration~\citep{venkatesh2019review}. Although some recent research work has proposed unsupervised feature selection~\citep{sheikhpour2017survey}, the lack of tag information limits those models' ability to capture the map knowledge between features and labels. Recent semi-supervised solutions, e.g., Semi-JMI, Semi-IAMB~\citep{sechidis2018simple}, still limit relations between features as Markov Blanket(MB)\citep{koller1996toward} or casual relations,e.g., Parents and Children relations(PC) \citep{yu2021unified}. Those simple relations can hardly express the complex latent structures existing among features. 

In recent years, self-supervised learning obtains supervisory signals from the data itself rather than the labels, often leveraging the underlying structure in the data~\cite{yoon2020vime}. Self-supervised learning has attracted great attention due to its incredible data efficiency and generalization ability, and many state-of-the-art models have followed this paradigm~\cite{li2017feature}, e.g., PatchNet\citep{yang2019patch}, GPT-3~\citep{brown2020language} and RIG\citep{nair2018visual}. It reduces the dependence on labels and can discover effective semantic or structural meanings with learnable intermediate representations. However, self-supervised learning has not been applied to general feature selection tasks, and the existing empirical self-supervised models are hard to design pretext tasks that can capture the complex structure in tabular data with "flat" features.

In order to tackle the problems mentioned above, this paper proposes an Attention-based Semi-supervised Feature Selection (A-SFS) in tabular data without the support of domain knowledge. This paper statements of contribution as follows:


\begin{enumerate}[]
\item \textbf{A new feature selection method based on self-supervised pattern discovery.}
This paper first proposes using self-supervised learning for the complex structure and relationship identification in tabular data. Rather than using predefined Graph theory or Markov Blanket~\citep{ling2019bamb} that needs custom-designed algorithms to generate special feature structures,  self-supervised learning empowers us to represent different structures with the neural network, which can express any form of the objective function.

\item \textbf{A multi-task self-supervised model for latent structure discovery.}
We designed a multi-task autoencoder to mine the underlying complex data structure valuable for feature selection from two different views. With the support of two pretext tasks, our design can automatically build a knowledge model to mine the different types of structure relationships for different tabular data, rather than artificial design.

\item  \textbf{Batch-attention based feature weight generation.}
The self-supervised learned structure knowledge is embedded into the batch attention model and effectively guides feature selection, by which only a small number of supervised samples are needed, and feature weights are generated in batch to reduce the impacts from misleading samples.

\end{enumerate}

Extensive experiments have been performed by comparing with ten state-of-art benchmarks, including LightGBM~\citep{ke2017lightgbm} and XGBoost~\citep{chen2016xgboost} on six representative real-world datasets. Results show that A-SFS outperforms the others in terms of feature selection accuracy and forecasting accuracy. Further analysis shows that it has a much lower demand for labeled data than other methods. On average, A-SFS only requires 9.7\%$\sim$21.4\% labels to achieve the baseline. Besides, A-SFS is robust to different types of noises and data missing. The detailed experimental results and source codes are attached in the supplementary materials. Of course, A-SFS normally demands a certain amount of unlabeled samples for pre-trains. However, due to the fast development of IoT devices, a large amount of unlabeled data usually are readily available from many applications\cite{goldman2000enhancing}.


The leftovers of the paper are organized systematically as follows. Section II pithily reviews the previous state-of-the-art work on feature selection and attention mechanisms related to our work. Section III introduces notations and preliminaries on semi-supervised feature selection. In Section IV, we describe the proposed Multi-task self-supervised framework and the A-SFS algorithm under this framework. Section V evaluates the effectiveness of the A-SFS algorithm in terms of both classification accuracy and noise robustness based on public datasets. Finally, we summarize this work and some future directions in Section VI.

\section{Related Work}

First of all, this section focus on various aspects of the cutting-edge feature selection works and attention mechanism.

\subsection{Feature Selection}
Feature selection is the process of discriminating the most informative, non-redundant, and relevant features to select a subset of the observed features when developing a predictive model, which draws massive attention for its reducing dimension  effectiveness.
The early supervised feature selection for flat data based on information entropy usually assumes that each feature is independent while ignoring their correlation and potential feature structure.
With supervised information driving, feature relevance usually is levied only via its correlation between feature and class targets\cite{guyon2003introduction}. 
Recent trends of feature selection methods are more focused on data with specific structures,e.g., graph~\citep{jiang2019joint}, tree~\citep{wang2015multi}, or streaming data~\citep{ding2014detecting}. They are more inclined to minimize the empirical error penalized by custom-designed structural regularization term. 



In comparison, deep-learning-based feature selection methods are considered to have the potential to cope with the "curse of dimensionality and volume" of big data thanks to its outstanding empirical performance. A major driver of progress for different datasets is reproducible DNN-based architectures that can efficiently encode raw data into meaningful representations easily implemented on new datasets and related tasks with high performance. 
Tabnet\cite{arik2021tabnet} builds on the availability of canonical DNN architectural codes while providing the key advantages of representation learning and end-to-end training of DNN-based approaches. Feature selection is committed during the self-supervised structural learning process. Thus, it is an unsupervised feature selection solution lacking supervised information guidance.  In the supervised DNN-based FS solutions, Li et al.\cite{li2016deep} firstly proposed a deep feature selection (DFS) by adding a sparse one-to-one linear layer. Those network weights are directly used as the feature weights, which might be sensitive to noise\cite{roy2015feature}. 
Another area of progress is using neural network models to estimate feature importance as a means of interpreting the models learned from propositional data. 
Our previous work AFS\cite{gui2019afs} firstly introduces the attention mechanism into the feature selection process.
SANs\cite{vskrlj2020feature} uses similar architecture as AFS for estimating feature importance. FIR\cite{wojtas2020feature} integrates a dual-network feature selection model architecture composed of selectors and operators to propose and evaluate the importance of all the features. But those DNN-based solutions highly rely on the information from labels and cannot work effectively in the dataset with limited labels.




Many researchers have proposed to use semi-supervised feature selection to mitigate label dependence problems using both unlabeled and labeled data for feature selection. 
The goal is to use labeled samples to maximize the separability between different categories and to use unlabeled samples to preserve the local data structure with graph theory~\citep{chang2014convex}, markov boundary~\citep{xiao2017gmdh} or relations between parents and children ~\citep{ang2015semi}. 
However, the existing semi-supervised feature selection solutions face significant challenges in learning the local structures due to limited-expression power for unsupervised feature learning, e.g., variance score, laplacian score, fisher score, and constraint score or limited to categorical features~\citep{sheikhpour2017survey}. 
Furthermore, they are not capable of handling large-scale data because of the time-consuming computation of the graph~\citep{chang2014convex}. 

In recent years, self-supervised learning has gained many research interests as it can discover very complex relations and structures in images, language, and videos. The neural network-based models can represent much more complex structures than predefined metrics. Thus, if we can take full advantage of self-supervised learning for latent structure discovery, it is possible to devise a powerful feature selection method for general tabular data.

\subsection{Attention Mechanism}
The attention mechanism can be viewed as the evaluation architecture of the features for feature importance that takes arguments and context information, focusing on the most pertinent information rather than all available information, conforming to the way of thinking of human information processing. 
It uses limited resources to quickly filter out high-value target areas from global information for the higher-level perceptual reasoning and the more complex information processing. It has been used in various fields such as visual images~\citep{mnih2014recurrent}, language translation~\citep{parikh2016decomposable}, and audio processing~\citep{chorowski2015attention} with spatial or temporal structures.

Attention focuses on structure's salient parts for input with spatial structure (such as pictures). 
For example, the RAM model~\citep{mnih2014recurrent} uses the attention mechanism to extract important picture location information to generate dynamic internal representations of the scene, achieving less pixel processing. 
Regarding the temporal structure of the signals(such as video), the attention mechanism obtains the current and previous input relationship through the RNN or LSTM stacked loop architecture, such as "Transformer" network architecture~\citep{vaswani2017attention} and STAT~\citep{yan2019stat}. 
Additionally, a non-local neural network using the self-attention approach has been proposed~\citep{wang2018non}.

The research discussed above usually provides domain-specific solutions based on attention mechanisms for the data with specific structures. However, these models are challenging to apply in tabular data lacking apparent spatial or temporal structure and prior knowledge. Our previous work, AFS~\citep{gui2019afs} first brought the attention mechanism to the feature selection domain for tabular data. However, it suffers performance degradation for the data with noises since it has no self-supervised learning mechanism to guide the selection process.

\section{Notations and Preliminaries}

\subsection{Notations and definitions}

In feature selection, we have a dataset $X$ with $N$ samples and $d$ dimensions (features), $\bm{X}=\left\{\bm{X}_l,\bm{X}_s\right\}\in \mathbb{R}^{d \times N}$, as the matrix of training data. According to the label availability, this dataset might consist of two subsets:$\bm{X}_l=\left\{ x_1,x_2,\ldots, x_l \right\}$ with corresponding label set $\bm{Y}_l=\left\{ y_1,y_2,\ldots, y_l \right\}$, and  $\bm{X}_s=\left\{ x_{l+1},x_{l+2}
,\ldots, x_{l+s}\right\}$, which is non-labeled. $l+s=N$. 

The goal of semi-supervised feature selection is to select a subset of features $\bm{s}\subseteq{1,2,\ldots,d}$, $|\bm{s}|=k$ and  $k<d$ from the $d$-dimensional features of the original data supported by a structure representation with $\mathbb{R}_s$, which is surrogate supervised information defined from unlabeled samples.
This function represents the intra-feature structures learned with the raw data $\bm{X}_s$ in the process of solving the surrogate task. The objective function is defined as follows:
\begin{equation}
  \min\mathbb{E} _{\left ( \bm{x}_l,{y}_l \right ),\left ( \bm{x}_s,{\hat{y}}_s \right ) }\left [\mathcal{L}^k \left (  \bm{x}_l,y_l,\mathbb{R}_s\left ( \bm{x}_s,\hat{y}_s \right )   \right )  \right     ] 
\label{eq:min}
\end{equation}


where the limited labeled samples $\left ( \bm{x}_l,{y}_l \right )$ and numerous unlabeled samples $\left ( \bm{x}_s,{\hat{y}}_s \right )$, respectively, obeys labeled distribution $P_{\bm{X}_l,\bm{Y}_l}$ and pseudo-labeled distribution $P_{\bm{X}_s,\bm{Y}_s}$, 
$\mathcal{L}(\cdot)$ is a global optimization loss function of the downstream feature selection task.



Here, we propose using a neural network that can be optimized by multi-task self-supervision to represent data structures. 
Each self-supervised network is trained on a large number of pseudo-label samples $\bm{x}_s\in\bm{X}_s$ to obtain a pre-trained model $\mathbb{R}_s$ by minimizing the self-supervised pretexts loss function $l_s$ as follows. 

\begin{equation}
   \mathbb{R}_s=min{\mathbb{E}_{\left(\bm{x}_{s,}\hat{y}_s\right)\sim {P_{\bm{X}_s,\bm{Y}_s}}}\sum_{t=1}^{T}\alpha_t}\left[l_s\left(\bm{x}_s,{\hat{y}}_s\right)\right]
\end{equation}
where the pseudo-label samples $\left(\bm{x}_{s},{\hat{y}}_s\right)$ follow the distribution $P_{\bm{X}_s,\bm{Y}_s}$,$T$ is the number of self-supervised tasks,
$\alpha_i$ is a pretext parameter to adjust each self-supervised task's contribution for feature selection.

\section{Architectural Design}
\begin{figure*}[thp]
\centering
\includegraphics[width=1.0\textwidth]{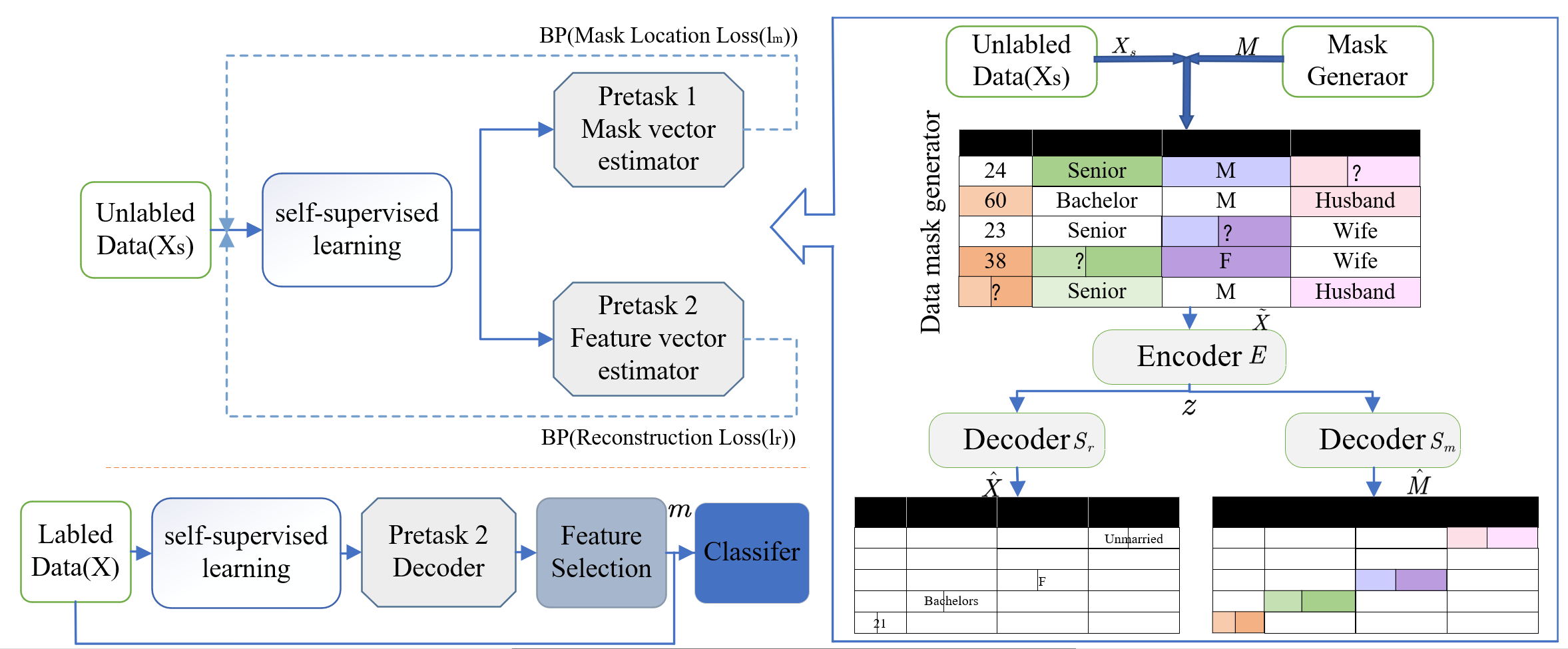}
\caption{The framework of feature selection model for tabular data based on multi-self supervision and batch-attention mechanism}
\label{fig:global_diagram}
\end{figure*}

Our proposed A-SFS architecture embeds a self-supervised model into the feature selection learning process similar to other embedded methods. Fig.~\ref{fig:global_diagram} illustrates the systematic self- and semi-supervised learning frameworks in A-SFS, consisting of two primary modules: the multi-task self-supervised module and the batch-wise feature selection module.

The multi-task self-supervised module is demonstrated in the upper-left part of Fig.~\ref{fig:global_diagram}, which consists of two pretext tasks responsible for obtaining structural information on the unlabeled data from multiple views.
The batch-wise feature selection module develops a batch-attention algorithm in supervised feature selection learning using the unique encoder network learned from the pretext tasks via multi self-supervised learning tasks, which aims to compute the weights for all features to select the optimal feature subset. The right of Fig.~\ref{fig:global_diagram} shows a novel masking method differing from the traditional masking, with two principal pretext tasks to learn the latent structure of tabular data in value, locations, and conjunction associated with features simultaneously.
 
\subsection{Data masking method of A-SFS}

Inspired by the denoising autoencoder (DAE), we propose an improved multi-task denoising autoencoder to learn the structure between data. 
Intuitively, the reconstruction loss of a simple encoding-decoding structure will assume that the pixels based on reconstruction are all independent, thereby reducing their ability to model correlations or complex structures.
As for A-SFS, a diverse set of tasks should lead to diverse feature combinations, more likely to cross the feature space required for general information understanding and fusion. Given the intuitive form of tabular data only rows and columns, we propose a novel masking method—Identical data masking.

The framework consists of a data mask generator $R$, encoder $E$, and decoder $D$. The data mask generator $R$ consists of a noise position matrix generator and a pretext generator. The noise position matrix generator outputs a binary mask matrix $\bm{M}$, whose output and unsupervised data $\bm{x}_s\in\bm{X}_s$ are used as the input of the pretext generator to generate the masked data for subsequent self-supervised processing.
The generation process in Eq.\ref{eq:maskmatrix},\ref{eq:masked_x}.

\begin{equation}
   \bm{M}=\left[\bm{m}_1,\bm{m}_2\cdots\bm{m}_d\right]\in\left\{0,1\right\}^d
   \label{eq:maskmatrix}
\end{equation}
\begin{equation}
  {\tilde{\bm{X}}}_s=R\left(\bm{x}_s,\bm{m}\right)=\bm{M}\odot{\bar{\bm{X}}}_s+\left(1-\bm{M}\right)\odot\bm{X}_s
  \label{eq:masked_x}
\end{equation}
where $\bm{m}_j$ obeys Bernoulli distribution with probability $p_m$. 
It means that the j-th feature of the i-th sample ${\bm{x}}_{i,j}\in {\bm{X}}_s$ has a probability ${\hat{p}}_{\bar{x}_{i,j}}=\frac{1}{N}\sum_{i=N_l+1}^{N_l+N_s}{\delta(\bar{x}_{i,j}=x_{i,j})}$ of being replaced by the j-th feature of other samples and transformed into ${\bar{\bm{x}}}_{i,j} \in {\bar{\bm{X}}_s}$.
Compared with the masking method of replacing with 0 or adding noise, $R\left(\bm{x}_s,\bm{m}\right)$ uses the same distribution data to replace the original data with a more indistinguishable double randomness. Explicitly, $\bm{m}$ is a random vector sampled from the Bernoulli distribution; Implicitly, ${\bar{\bm{x}}}_j$ is replaced by a random sample of $\bm{x}_j$, which ensures that the masked data ${\tilde{\bm{X}}}_s$ not only retains the tabular structure of the original data $\bm{X}_s$, but also maintains the same distribution, making it more difficult to distinguish from origin. 


\subsection{Multi-task self-supervision for structure discovery}
Following the most widely used autoencoder, an encoder $E:{\tilde{\bm{X}}}_s\rightarrow \bm{z}$ creates a low-dimensional vector $\bm{z}$ containing the meaning of the input data in a hidden layer $\bm{z}=E\left(\bm{x}_s \right)$. Then there are pretext decoder models $D:\bm{z} \rightarrow \hat{y}_s$ training jointly that reconstruct the input data or other surrogate data from the informative representation $\bm{z}$ to solve pretext tasks defined with pseudo-label $\hat{y}_s \in \bm{Y}_s$.


We aimed for tasks that were conceptually simple yet as diverse as possible, and these multi-tasks are inter-correlated. Normally, learning these tasks jointly could improve performance compared to learning them individually and seek each task's balance in the optimization process.
After encoding, there is no difference between different features with the same value, which leads us to overlook the physical meaning of the feature often. 
For example, although \textit{education}, \textit{race}, and \textit{occupation} are all encoded between 0 to 1, the connection between \textit{education} and \textit{occupation} is much closer than \textit{race} and \textit{occupation}. We can predict the \textit{education} by a person's \textit{occupation}.
In order to solve this problem, we designed two pretext tasks based on the characteristics of the tabular data:

\begin{enumerate}
    \item \textit{Location estimation:} predict mask vector; 
    \item  \textit{Feature estimation:} predict masked values.
\end{enumerate}

In the absence of domain knowledge, we can use the location information of features as category information to distinguish differences in the physical meaning.
An uncultivated and rough combination of self-supervision tasks might cause input channel conflict and learning tasks conflict, impeding performance. We introduce a shared encoder and a separate decoder network for each pretext task.
Both decoder work on the representation $\bm{z}$ output by the shared encoder $E$, trying to evaluate the mask matrix $\bm{m}$ and the masked data $\bm{x}_s$ collaboratively.
The appealing feature of joint feature selection across a group of related tasks is that it encourages multiple predictors to share similar embedding patterns and shape them. 
The two models and their functions are as follow:

\noindent\textbf{Location estimation.}$S_m:{\bm{z}}\rightarrow \hat{\bm{m}}\in\left\{0,1\right\}^d$, input the representation vector $\bm{z}$ and output which feature is masked by counterpart.

\noindent\textbf{Feature estimation.}$S_r:{\bm{z}}\rightarrow{\hat{\bm{x}}_s}$, input the representation vector $\bm{z}$ and output the estimated value of the original data.

The shared encoder $E$ and the separated multiple decoders $D$($S_m$ and $S_r$) are jointly trained for the following objective functions:
\begin{equation}
    \min_{E,S_{m},S_{r}}\mathbb{E}_{\bm{x}\sim p_{\bm{x}},\bm{m}\sim p_{\bm{m}},\widetilde{\bm{x}}\sim R\left (\bm{x},\bm{m} \right )}\left [l_{m}\left ( \bm{m},\hat{\bm{m}}  \right )    +\alpha l_{r}\left ( \bm{x}_{s},\hat{\bm{x }}_s \right )  \right ]
\end{equation}
where $\hat{\bm{m}}=(S_m\circ E)({\tilde{\bm{x}}}_s)$,${\hat{\bm{x}}}_s=(S_r\circ E)({\tilde{\bm{x}}}_s)$.The first loss function $l_{\bm{m}}$ is the sum of the binomial cross-entropy loss of the mask matrix:
\begin{equation}
\begin{aligned}
        l_{\bm{m}} \left(\bm{m},\hat{\bm{m}}\right)= -\frac{1}{d} \left [ \sum_{j=1}^{d}m_j\log_{2}{\left [\left (S_m \circ E\right )_j \left (\widetilde{\bm{x}}_s\right)\right]}\right]+\\
      -\frac{1}{d} \left [\left(1-m_j\right)log_{2}{ \left[1-\left(S_m \circ E \right)_j\left (\widetilde{\bm{x}}_s \right )\right ]} \right ] 
\end{aligned}
\vspace{-0.08in}
\end{equation}
The second loss function $l_r$ is the reconstruction loss:
\begin{equation}
    l_r\left(\bm{x}_s,{\hat{\bm{x}}}_s\right)=\frac{1}{d}\left[\sum_{j=1}^{d}\left(\bm{x}_j-\left(S_r\circ E\right)_j\left(\widetilde{\bm{x}}_s\right)\right)^2\right]
\end{equation}
$\alpha$ controls the proportion of the two concern losses.

Structure relationships imply the underlying mechanism about the data and class variable, and thus they are persistent across different settings or environments.
The two self-supervised tasks we designed complement each other, the feature estimation task ensures the integrity of the original information, and the location estimation task ensures the correlation between the features. $S_m$ infers the mask position from the inconsistency of the feature value, and $S_r$ derives the original value from the relevant non-masked  features. The entire process does not require hand-designed criteria and domain-knowledge priors so that the network can be applied in any self-supervised structure of tabular data without an explicit structural performance like images or language.

\subsection{Batch-attention feature weight generation}

This paper implements a feature selection algorithm for joint training with a deep network. 
We convert the correlation mapping between the features and the supervision target into the probability problem of the feature being selected or not. In the supervision problem, whether features should be selected depends on the distribution of weights in the feature selection pattern.
In this stage, the attention mechanism is utilized to sort the features according to the criteria of redundancy minimization among the feature subset and relevance maximization between the features in the subset and their labels.

The batch-attention feature selection module consists of two sub-modules: i) Batch-wise weight generation; ii) Evaluation module.
\subsubsection{Batch-wise weight generation}
The weight generation module is performed in three steps. First, it performs an attention transformation network to compress the data sample into a low-dimensional information representation, and then maps it back to an instance-wise weight matrix of the same size as the input. 
\begin{equation}
     \bm{\tau}_i =\bm{W}_2\cdot\tanh{\left(\bm{W}_1\cdot\left(S_r\circ E\left(\bm{x}_i\right)\right)+\bm{b}_1\right)}+\bm{b}_2
\end{equation}
where $\bm{W}_1\in\bm{R}^{B\times d},\bm{W}_2\in\bm{R}^{d\times B},\bm{b}_1\in\bm{R}^B,\bm{b}_2\in\bm{R}^d$. 
It should be noted that the attention transformation network is generic and portable. Here we stack two dense layers to realize it. 

In the second step, we explicitly average the $\bm{\tau}$ obtained in the previous step in minibatch to obtain a fixed score vector $\bar{\bm{\tau}}$.
Generally speaking, an input sample corresponds to a weight mapping after the attention transformation module processing, but in feature selection, it is generally assumed that all given samples have the same significant feature even individual noisy data, so it is not desirable to use instance-wise feature selection, while batch-wise feature selection can better eliminate the influence of special distribution noise. So for each minibatch $B$:
\begin{equation}
    \bar{\bm{\tau}}=\frac{1}{B}\sum_{i=1}^{B}\bm{\tau}_i
\end{equation}

In the last step, it normalizes the selecting probability $\bar{\bm{\tau}}$ by softmax to generate differentiable results between 0 and 1.
\begin{equation}
    \bm{a}=softmax\left(\bar{\bm{\tau}}\right),with,          {\bm{a}}_i=\frac{e^{{\bar{\bm{\tau}}}_i}}{\sum_{j=1}^{d}e^{{\bar{\bm{\tau}}}_j}}
\end{equation}


\subsubsection{Evaluation module}

The hierarchical batch-attention structure reduces the unique particular distribution data's tricky to the feature importance evaluation.
In order to continuously adjust the attention matrix according to the optimization goal of the task so that it can accurately express the correlation between the feature and the task goal, an evaluation module is designed. Its core is to make a trade-off between selection and de-selection by solving the objective function for backpropagation, and continuously adjust the weight until convergence.
The supervised feature vector $\bm{X}_l$ and the weight $\bm{a}$ are multiplied in pair-wise $\odot$ to obtain the weighted feature $G=\bm{X}_l\odot\bm{a}$, 
The evaluation module needs to input the weighted feature $G$ into the classifier and calculate the difference between the excitation response and the target output corresponding to the training input. The obtained response error is iteratively updated according to the following loss function to evaluate the weight value's correctness better.

 \vspace{-0.1in}
\begin{equation}
    \mathop {argminloss}_{E,S_r,S_m,\theta_f} \left[f\left(S_r\circ E\left(x_l\right)\right)-y_l\right]
    \label{eq:loss_func}
\end{equation}
Where $f(\cdot)$ is a batch-attention feature selection network with parameters $\theta_f$, the loss function of the evaluation network $\left(\ref{eq:loss_func}\right)$ depends on the learning criteria and the type of optimization objective.
The regression loss uses the mean square error(MSE), and the classification tasks uses the categorical cross-entropy loss.
It is worth noting that since the attention weight generation module and evaluation module are loosely connected and can be trained separately.
$f(\cdot)$ is replaceable and reusable.
For a specific learning problem, $f(\cdot)$ can be replaced accordingly with the most appropriate network structure, such as DNN, CNN, and RNN.
For many evaluation tasks, excellent specialized models such as ResNet and VGG can be transferred. This is achieved by initializing the evaluation module parameters of A-SFS to the weights of the exposed model.

\section{Experiments}
\label{sec:experiment}
In this section, we introduce the experimental comparison method, dataset, evaluation metrics, and settings.

\subsection{Comparison method}
\label{sec:Comparison method}
  In order to evaluate the effectiveness of our proposed method, A-SFS is compared with eleven wildly used strong baselines, including eight supervised methods, two semi-supervised methods and one unsupervised methods.

\noindent\textbf{Supervised methods:}\footnote{http://featureselection.asu.edu}

Traditional Supervised Feature Selection Models:
\vspace{-0.1in}
\begin{itemize}[]
    \setlength{\itemsep}{0pt}
    \setlength{\parsep}{0pt}
    \setlength{\parskip}{0pt}
    \item \textit{Fisher Score:}(FSCORE) similarity-based methods, which selects each feature independently according to their scores under the fisher criterion; 
    \item \textit{$ll-l_{21}$:} sparse-learning-based methods, which consider smooth convex optimization via Nesterov’s method~\citep{liu2012multi}; 
    \item \textit{CIFE:} information-theoretical-based methods, which maximizes the joint class-relevant information by reducing the class-relevant redundancies among features~\citep{lin2006conditional};
    \item \textit{Trace Ratio:}(TR) similarity-based methods, which select optimal feature subset based on the corresponding score on the subset-level score~\cite{nie2008trace};
\end{itemize}
\vspace{-0.1in}

Tree-Based Supervised Feature Selection Model:
\vspace{-0.1in}
\begin{itemize}[]
    \setlength{\itemsep}{0pt}
    \setlength{\parsep}{0pt}
    \setlength{\parskip}{0pt}
    \item \textit{Random\_Forest:}(RF) a tree-based feature selection method provided by the scikit-learn package, which operates by constructing a multitude of decision trees at training time; 
    \item \textit{LightGBM:}(LGB) a highly efficient gradient boosting framework that uses tree-based learning algorithms;
    \item \textit{XGBoost:}(XGB) a scalable tree gradient boosting system;
\end{itemize}
\vspace{-0.1in}

DNN-Based Supervised Feature Selection Model: 
\vspace{-0.1in}
\begin{itemize}[]
    \setlength{\itemsep}{0pt}
    \setlength{\parsep}{0pt}
    \setlength{\parskip}{0pt}
    \item \textit{AFS:} an attention-based feature selection via  a shallow attention net for each feature and a learning module;
    \item \textit{FIR:} a dual-net feature selection model architecture consisting of operators and selectors based on a stochastic local search process\cite{wojtas2020feature};
    \item \textit{SANs:} a feature importance exploration model using representation vector based on single-layer self-attention network architecture\cite{vskrlj2020feature};
    \item \textit{Tabnet:} a canonical deep tabular data learning architecture using sequential attention to learn a ‘decision-tree-like’ mapping\cite{arik2021tabnet}.
\end{itemize}

\vspace{-0.1in}
\noindent\textbf{Semi-supervised method}~\citep{sechidis2018simple}:
\vspace{-0.1in}
\begin{itemize}
    \setlength{\itemsep}{0pt}
    \setlength{\parsep}{0pt}
    \setlength{\parskip}{0pt}
 \item  \textit{Semi\_JMI:} combining theoretical and empirical studies with some “soft” prior knowledge of the domain, which takes into account relevancy and redundancy; 
 \item  \textit{Semi\_MIM:} semi\_supervision with information-theoretic methods which ranks derived through MIM criterion; 
\end{itemize}

\vspace{-0.1in}
\noindent\textbf{Unsupervised method: }
\vspace{-0.1in}
\begin{itemize}
\item \textit{UDFS:} Unsupervised Discriminative Feature Selection method that selects features by exploiting both the discriminative information and feature correlations~\citep{yang2011l2}.
\end{itemize}

\subsection{Datasets}
\label{sec:Dataset}
We conduct experiments on 6 datasets from the repository of the Open Machine Learning project\footnote{https://www.openml.org/}.
The smallest dataset (Pendigits) has 16 features and 1000 samples to train feature weight, while the largest dataset (MNIST and Fashion MNIST(FMNIST)) has 784 features and 10000 samples.
The datasets are fully observed; The attribute values are scaled to [0,1] with a MinMax scaler.
We use unsupervised train data(Train(unlabeled) in Table~\ref{tab:datasets}) in the self-supervised module and only use limited supervised train data(Train(label) in Table~\ref{tab:datasets}) in the feature selection module and select TOP K features for downstream classification.

\begin{table}[h]
 \centering
  \caption{Datasets Properties}
    \begin{tabular}{lrrrrr}
    \toprule
    \textbf{Dataset}  & \textbf{Feat.} &  \textbf{Labeled/un} & \textbf{Test} & \textbf{Class} & \textbf{TopK} \\
    \midrule
    MNIST &784 & 3000//10000 & 10000 & 10&30 \\
    FMNIST &784   &  3000//10000 &  10000 &  10 &   30\\
    USPS &256 & 3000//9000& 3000& 10&30\\
    Auml\_url  & 80  & 3000//10000  &  3000  & 5 & 5\\
    Pendigits &16 &1000//7694 & 1000& 10& 5\\
    Puish\_url & 80 &3000//10000 & 3000& 5& 5\\
    Optdigits  &  64 & 1000//5620 & 1000 & 10 & 5\\
    \bottomrule
    \end{tabular}
    \label{tab:datasets}
\end{table}%

\subsection{Settings and evaluation metrics}
\label{sec:Settings and evaluation metrics}
  
We evaluate the A-SFS method and benchmarks on six public datasets, evaluate the feature subset by the LightGBM classifier and choose the test set's F-1 score to measure the model effect. 

    \noindent\textbf{A-SFS configurations}: For the self-supervised pretext, we adopt a 2-layer fully-connected layer with hidden units and sigmoid activation as the encoding layer network, and the decoding network adopts the same settings.
All conducted multi-task self-supervised experiments use the RMSprop optimizer to train the self-supervised module of A-SFS for 40 epochs with a learning rate of 0.001.
Among them, the self-supervision coefficient is adjusted to $\alpha=2$, and the mask matrix probability $p_m=0.2$. The batch sample is 128.
    We modify the eigenvector estimation task to mean squared error and the mask vector estimation task to binary cross-entropy loss.
As for the batch-attention feature selection module is designed to involve a 2-layer fully connected layer with 300 hidden units and tanh activation to present reasonable weights.
All methods are compared using an evaluation network $f(\cdot)$  with the same architecture for fair, involving 3 fully connected layers and RELU activations.
The feature selection stage was trained for 2000 epochs using the Adam optimizer\citep{kingma2014adam} with a learning rate of 0.001.

\noindent\textbf{General settings}: 
All related hyperparameters of the methods involved adopt the default settings.
Each method has its own feature importance sequence sorted by numerical value fits with the top k features in the test set according to the ranking, and puts them into the benchmark classifier to evaluate the effect of its feature subset. We report the average evaluation results of all methods on the test set.

\begin{table}[htp]
  \begin{center}
  \caption{Downstream classification accuracy based on top 5/30 selected features on different datasets with 14 feature selection methods, using 5 times of 5-fold cross-validation to provide a fair comparison.}
  \label{tab:acc}
  \resizebox{1\textwidth}{!}{
  \renewcommand\arraystretch{1.2}
    \begin{tabular}{cccc cccc}
    \toprule
    Name &MNIST & FMNIST &USPS &Auml\_url &Pendigits &Puish\_url &Optdigits  \\
    \midrule
     LGB &80.05$\pm$.22 &   80.72$\pm$.25 &92.83$\pm$.21    & \textbf{89.95$\pm$.35} & 84.25$\pm$.26 &\textbf{89.11$\pm$.16} & 	\underline{73.54$\pm$.21}  \\
    \cline{1-8}
    XGB& 76.88$\pm$.34  & 80.27$\pm$.03  & \underline{93.15$\pm$.28}    & 76.09$\pm$.34 & 86.73$\pm$.73 & 76.69$\pm$.30 & 70.60$\pm$.31    \\
    \cline{1-8}
    RF      &81.06$\pm$.33 &73.99$\pm$.11   &93.24$\pm$.19    &85.76$\pm$.24  &  85.86$\pm$.34&	86.94$\pm$.28	&72.27$\pm$.37	\\
    \cline{1-8}
    AFS     & \underline{81.80$\pm$.69}&    69.84$\pm$.21	  &91.81$\pm$.56	&84.65$\pm$.57&	85.46$\pm$.51&	85.76$\pm$.55	&73.51$\pm$.63  \\
    \cline{1-8}
    FIR     &71.40$\pm$.35 & \underline{81.24$\pm$.05}&	92.54$\pm$.39 &	83.42$\pm$.32 &	85.43$\pm$.41 &	80.13$\pm$.44 &	40.36$\pm$.29  \\
    \cline{1-8}
    SANs&	70.69$\pm$.15 & 78.98$\pm$.04&	91.84 $\pm$.25	&76.19$\pm$.31 &	\underline{87.66$\pm$.33} &	81.10$\pm$.43& 	26.65$\pm$.35 \\
    
    \cline{1-8}
    Tabnet    &	72.83$\pm$.54&77.62$\pm$.02 &	92.58 $\pm$.03	&77.15$\pm$.78 &	87.63$\pm$.01 &	84.71$\pm$.04& 	60.81$\pm$.02  \\
    
    \cline{1-8}
    FSCORE &70.50$\pm$.79 &   65.43  $\pm$.81  &85.64$\pm$1.64 &	69.99$\pm$1.16&	86.56$\pm$1.08&	 73.61$\pm$1.27&	69.54$\pm$1.58\\
    \cline{1-8}
    Ll\_l21 &	46.70$\pm$.49&  63.78  $\pm$.45&	92.97$\pm$.24&	83.71$\pm$.52&	82.70$\pm$.59&	83.71$\pm$.62&	54.07$\pm$.73\\
    \cline{1-8}
    CIFE &	70.75$\pm$.38&  57.03  $\pm$.32&	82.44$\pm$.66&	85.69$\pm$.06&	86.34$\pm$.10&	84.29$\pm$.25&	72.23$\pm$1.49\\
    \cline{1-8}
    TR	&69.32$\pm$.46&   51.33$\pm$.63&	85.60$\pm$.23&	63.15$\pm$1.06&	86.42$\pm$.17&	53.61$\pm$1.28&	46.29$\pm$.69\\
    \cline{1-8}
    UDFS &	77.18$\pm$.52&  72.53$\pm$.72&	90.26$\pm$.57	&73.12$\pm$.47&	77.45$\pm$.41&	83.20$\pm$.45&	51.93$\pm$.45\\
    \cline{1-8}
    Semi\_JMI &	69.46$\pm$.12&  72.44$\pm$.07    &	87.94$\pm$.08&	87.51$\pm$.11	&86.20$\pm$.59&	87.73$\pm$.08&	60.25$\pm$.08
    \\
    \cline{1-8}
    Semi\_MIM &	67.95$\pm$.63&   70.83$\pm$.15  &	87.19$\pm$.13&	80.23$\pm$.21	&85.55$\pm$.50	&86.61$\pm$.18	&58.96$\pm$.44	\\
    \cline{1-8}
    A-SFS &	\textbf{83.75$\pm$.25}&   \textbf{82.48  $\pm$ .03} &	\textbf{93.87$\pm$.15}&	\underline{89.65$\pm$.22}&	\textbf{88.52$\pm$.18}&	\underline{88.10$\pm$.25}&	\textbf{75.47$\pm$.22}	\\
    \bottomrule
    \end{tabular}
    }
    \end{center}
\vspace{-2em}
\end{table}

 \subsection{Performance comparisons}
Table~\ref{tab:acc} shows the average classification result with standard deviation using the top-5/30 features selected by the benchmarks and  A-SFS methods. As can be seen from this table, A-SFS significantly outperforms the compared methods on most of the compared datasets. It achieves about 1.95\%$\sim$13.25\%,1.93\%$\sim$23.54\% absolute accuracy improvements on the MNIST and Optdigits datasets, respectively, compared with other benchmarks.

\begin{figure}[thp]
    \centering
    \setlength{\abovecaptionskip}{0pt}
    \setlength{\belowcaptionskip}{0pt}
    \begin{subfigure}[t]{0.49\linewidth}
        \begin{minipage}[b]{1\linewidth}
        \centering
        \includegraphics[width=1\linewidth]{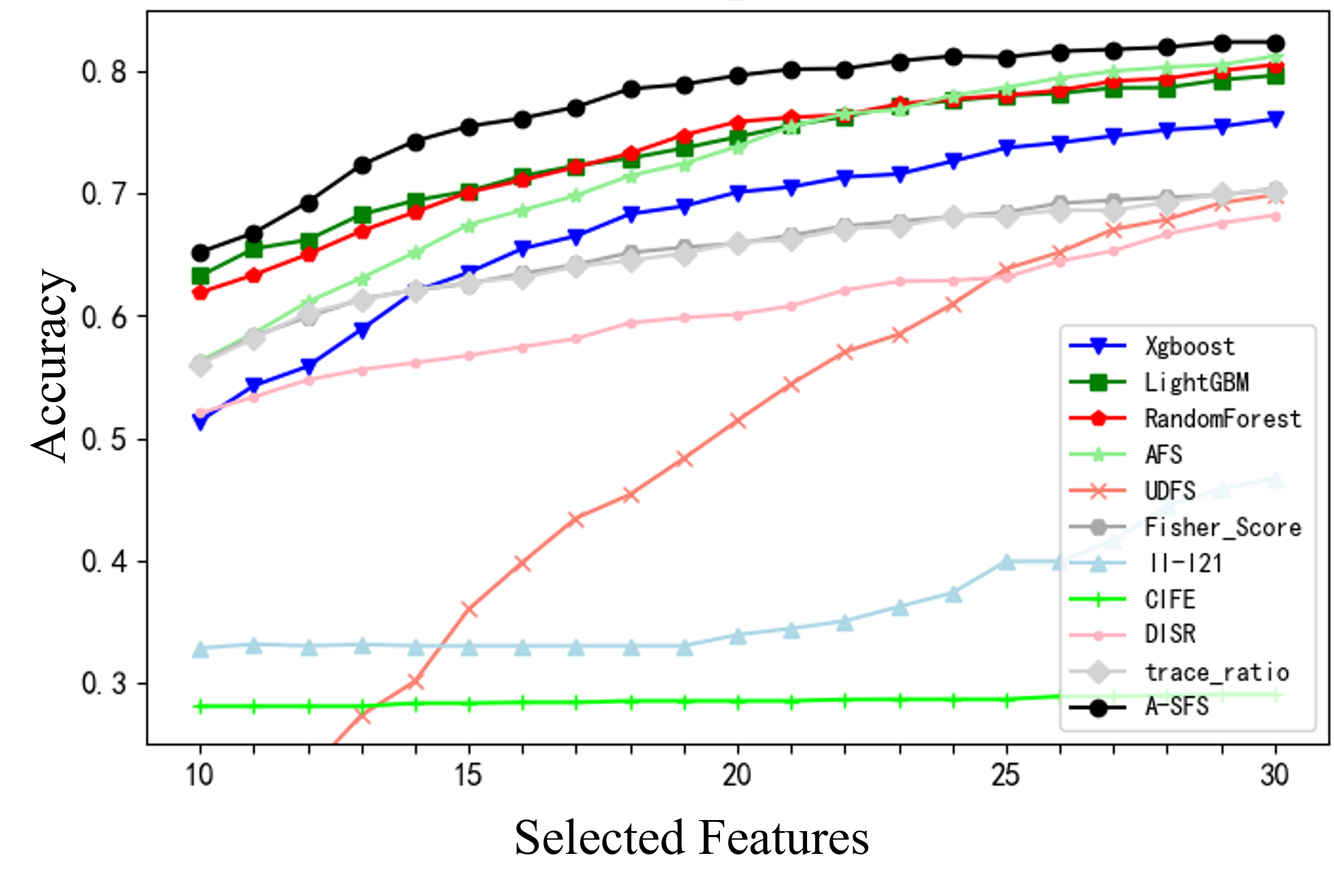}
        \caption{MNIST\_Gaussian}
        \label{a}
        \end{minipage}
    \end{subfigure}
    \begin{subfigure}[t]{0.49\linewidth}
        \begin{minipage}[b]{1\linewidth}
        \includegraphics[width=1\linewidth]{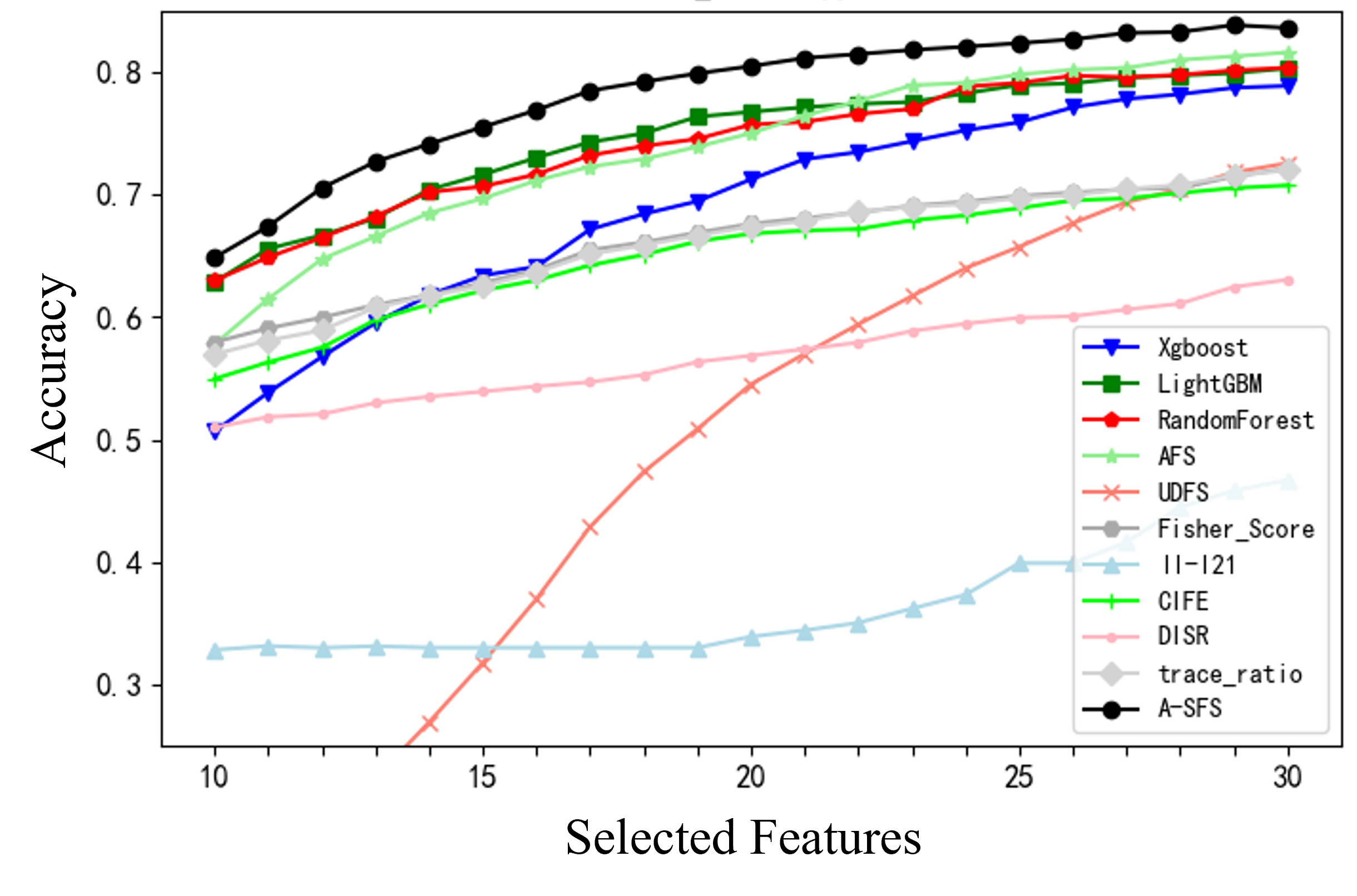}
        \caption{MNIST\_Salt\&Pepper}
        \label{b}
        \end{minipage}
    \end{subfigure}
    \begin{subfigure}[t]{0.49\linewidth}
        \begin{minipage}[b]{1\linewidth}
        \includegraphics[width=1\linewidth]{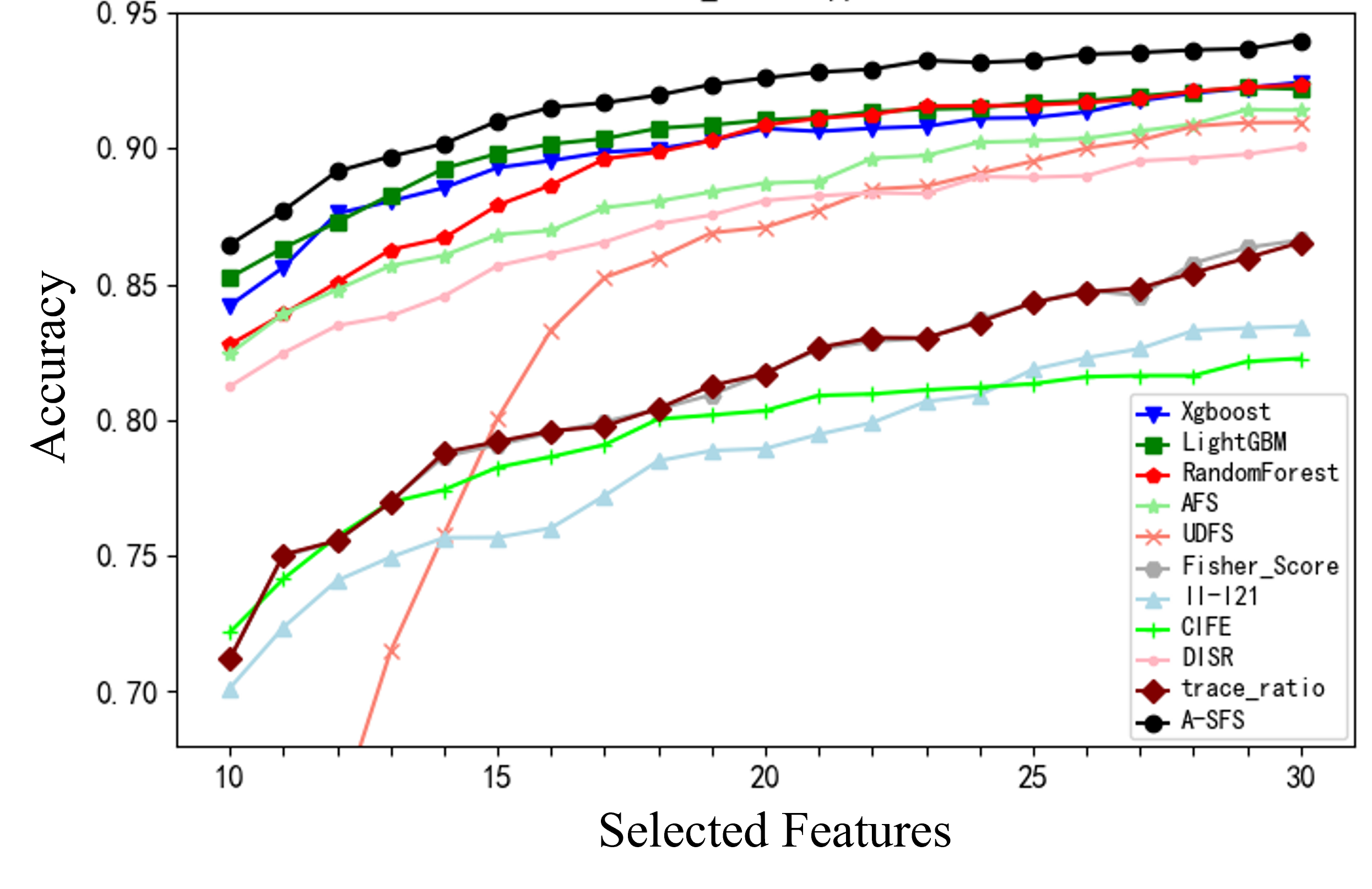}
        \caption{MNIST\_Speckle}
        \label{c}
        \end{minipage}
    \end{subfigure}
    \begin{subfigure}[t]{0.49\linewidth}
        \begin{minipage}[b]{1\linewidth}
        \includegraphics[width=1\linewidth]{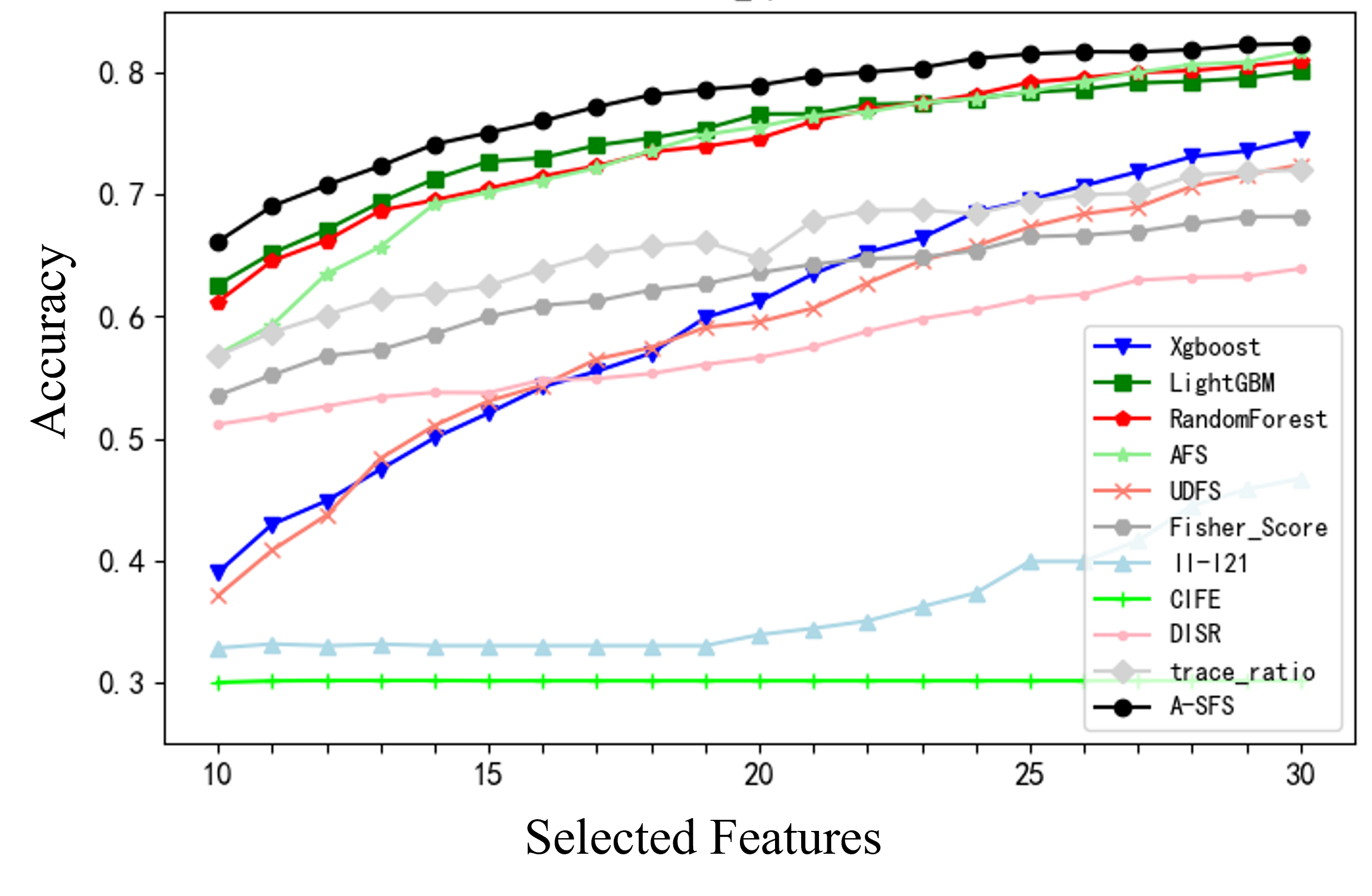}
        \caption{USPS\_Gaussian}
        \label{d}
        \end{minipage}
    \end{subfigure}
    \begin{subfigure}[t]{0.49\linewidth}
        \begin{minipage}[b]{1\linewidth}
        \includegraphics[width=1\linewidth]{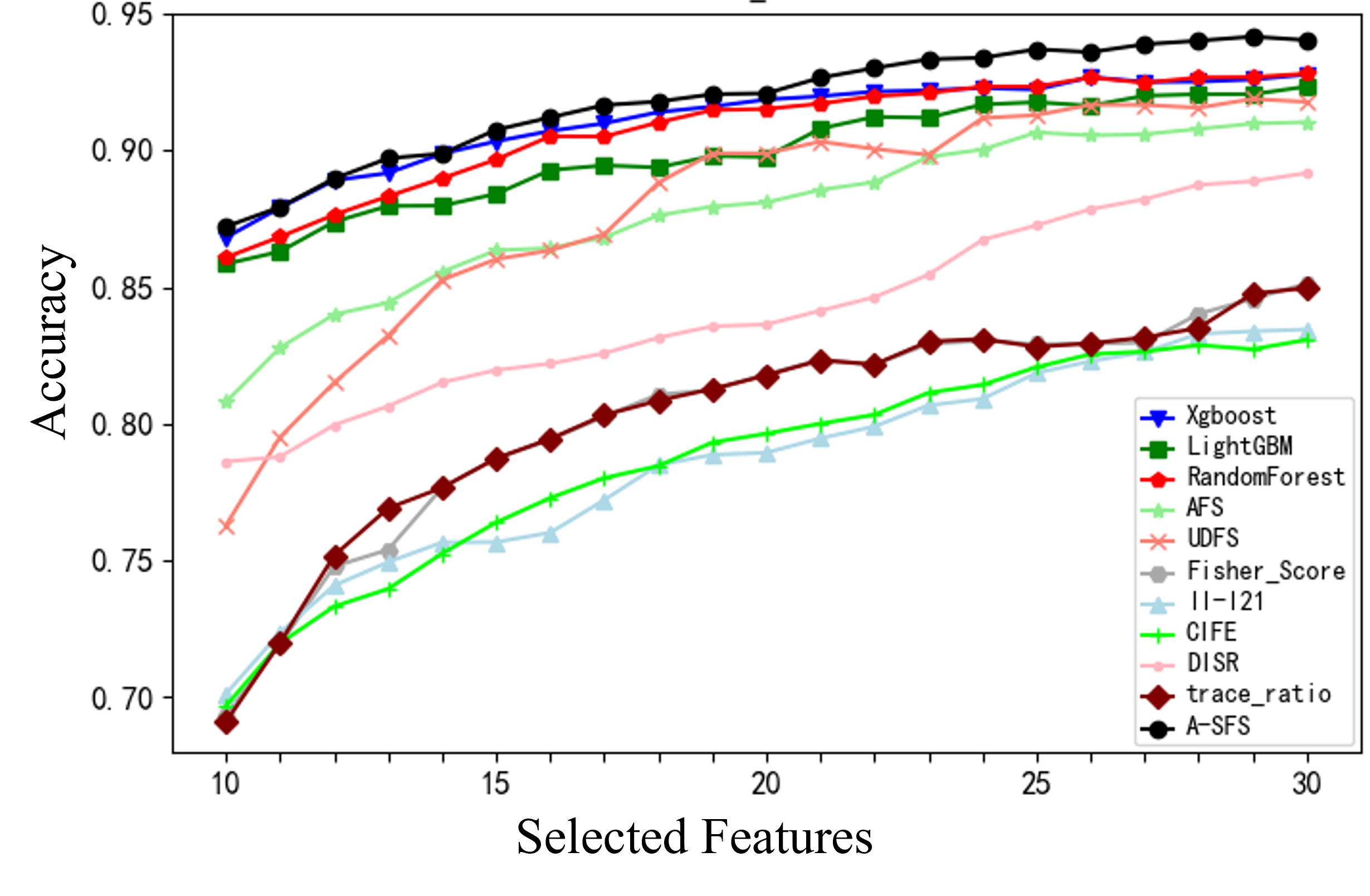}
        \caption{USPS\_Salt\&Pepper}
        \label{e}
        \end{minipage}
    \end{subfigure}
    \begin{subfigure}[t]{0.49\linewidth}
        \begin{minipage}[b]{1\linewidth}
        \includegraphics[width=1\linewidth]{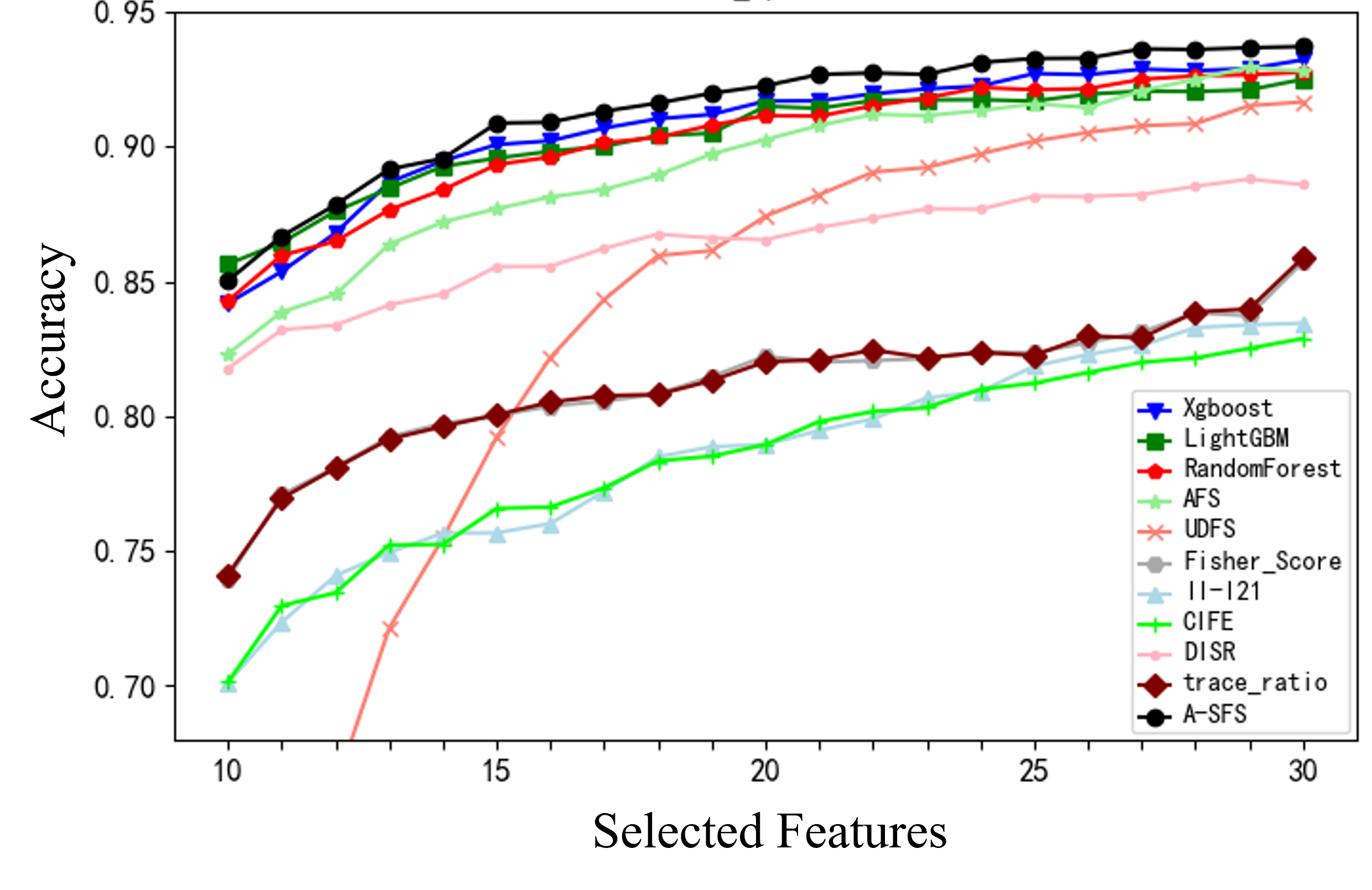}
        \caption{USPS\_Speckle}
        \label{f}
        \end{minipage}
    \end{subfigure} 
    \caption{Accuracy comparisons among 11 feature selection methods in MNIST, USPS datasets when adding Gaussian, S\&P, Speckle noise. The numbers of selected features are set from 10-30, results for A-SFS are after 2000 training steps.}
\label{fig:graph}
\end{figure}

Conventional methods, e.g., FSCORE, $ll-l_{21}$, CIFE and TR, partially due to their simple metrics and lack of feature structure guidelines, typically have poor performance on those datasets. The decision tree-based solutions: RF, LGB and XGB, normally achieve comparable good performance due to their strong feature identification during decision tree construction. However, their performance is not stable in different datasets, and they are sensitive to the noise, especially XGB. In the Auml\_url and Puish\_url, XGB suffers significant performance deterioration while LGB has very good performance on those two. For the DNN-based solution AFS, due to the lack of feature structure discovery, its performance is not comparable with A-SFS. FIR and SANs highly rely on the information from labels and cannot work effectively in the dataset with limited labels.


The semi-supervised and unsupervised solutions display generally mediocre performance. Semi\_JMI and Semi\_MIM rely on causality-based graph generation. The optimal surrogate to be used in ranking the features is strongly dependent on soft prior knowledge.
For those datasets, it is hard to build those casual graphs, and they have a good performance on small datasets while poor performance on comparable bigger datasets, e.g., MNIST and USPS. The unsupervised solution, TabNet and UDFS have poor performance in most datasets as it lacks label guidance. A-SFS outperforms TabNet with 15.0\% in MNIST and 24.1\% in Optdigits.


These results reveal that although general tabular data lack explicit spatial correlation and semantic connection, A-SFS can still mine the implicit structural relations between features to maintain structural flow and then take advantage of unlabeled tabular datasets to enhance the robustness of the model and the ability to select important features.







\subsection{Noise robustness analysis}

This section analyzes the classification performance and labels dependence of A-SFS and benchmarks under three different types of noises: Gaussian, Salt\&Pepper and Speckle. The implementation of noise comes from the skimage library\footnote{https://scikit-image.org/docs/dev/api/skimage.util.html}. 

\noindent  \textbf{Performance analysis under noises}:

In Fig.~\ref{fig:graph} the Gaussian noise $\delta\sim{N(0,0.01)}$. The Salt\&Pepper Noise has high grayscale noise and low grayscale noise simultaneously with a ratio of 0.5. The Specle noise is $out=data+n\times data$ where n is Gaussian noise $n\sim{N(0,0.3)}$. 

As shown in Fig.~\ref{fig:graph}(a)(b)(c) for MNIST and Fig.~\ref{fig:graph}(d)(e)(f) for USPS, A-SFS still achieves the best accuracy in all ranges for different datasets under three types of noises. Its performance is significantly better than all the compared ten methods. A-SFS has an absolute accuracy improvement of $2\%\sim13\%$ compared to LGB and XGB in the MNIST dataset. Compared with AFS and UDFS, there are more obvious advantages. The same effect also appears in the USPS datasets.  

LGB and XGB are boosting methods based on iterative algorithms. Each iteration reduces the deviation of the model. However, they are generally sensitive to noise. The DNN-based AFS also suffers from the misleading of noises. Thus, their performance deteriorates when there are noises exists compared to results on the dataset without noises. In comparison, A-SFS can effectively mitigate the impacts introduced from various noises as it can use the feature structure that existed among features.

\noindent\textbf{Visualization of noise robustness}:

To obtain a more intuitive understanding of the superiority of the proposed A-SFS method, we exhibit the learned feature visualization further to analyze the effectiveness of the two pretext tasks. We add six types of noises with a default noise ratio of 0.05. The settings of those noises are listed as follows:


\begin{itemize}
    \setlength{\itemsep}{0pt}
    \setlength{\parsep}{0pt}
    \setlength{\parskip}{0pt}
    \item Gaussian Noise: Gaussian-distributed additive noise, default setting: mean=0 and var =0.01;
    \item Salt \& Pepper Noise: High grayscale noise and low grayscale noise appear simultaneously, showing black and white noise. The ratio of high grayscale and low grayscale noise is 0.1;
    \item Poisson Noise: Poisson-distributed noise generated from the data;
    \item Speckle Noise:$out=data+n\times data$ where n is Gaussian noise with specified mean and variance, we set mean=0, var=0.3;
    \item Gaussian Blur: The data is convolved with a Gaussian function. In this process, high-frequency signals are filtered, and low-frequency signals are retained;
    \item Mean Blur: the output pixel value of the filter used in the smoothing process is the mean value of the pixel value in the kernel window (all pixels have the same weighting coefficient).
\end{itemize}

\begin{figure*}[htp]
\centering
    \includegraphics[width=1.\textwidth]{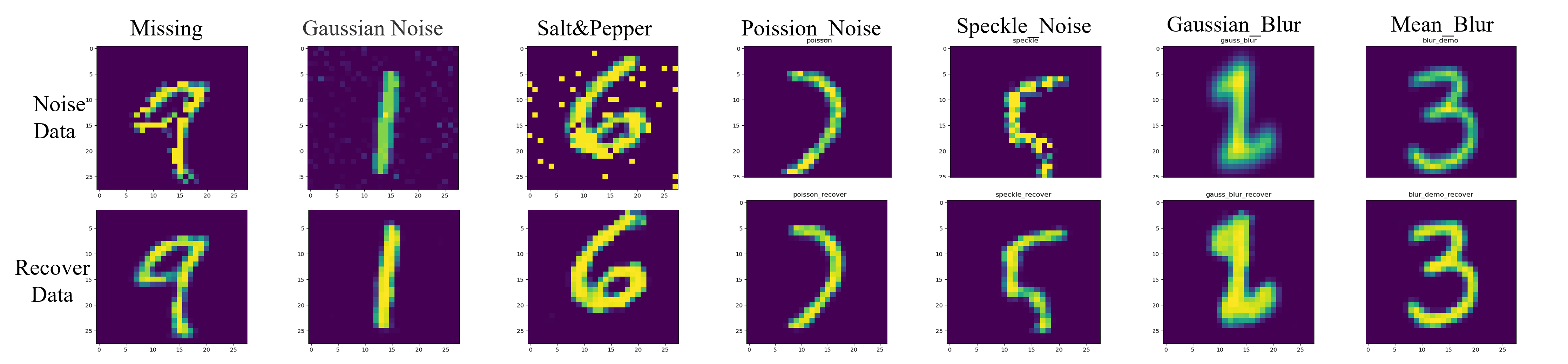}
    \caption{Pre-processing results of the proposed self-supervised module on the images with different noise.}
    \label{fig:my_visualize}
\end{figure*}

It is observed from Fig.~\ref{fig:my_visualize} that the self-supervision structure of A-SFS is robust against random noise. It can realize the location and restoration of noise or abnormal data. The reason is that the designed self-supervised structure can automatically construct a series of hidden feature knowledge, such as correlation information, structural relationship, importance, etc.
Knowing these features helps $S_m$ the decoding process locate the replaced noise data through the non-predictive performance between the feature values. For instance, if the value of a feature is very different from its correlated features, this feature is likely masked and corrupted. Then,  $S_r$ calculates the actual value of the replaced noise data through the relevant non-noise features.
In this way, the instability and perturbation issues of most single feature selection algorithms can be alleviated, and the subsequent learning tasks can be enhanced.

\subsection{Label dependence}

\begin{figure}[thb]
    \centering
    \setlength{\abovecaptionskip}{0pt}
    \setlength{\belowcaptionskip}{0pt}
    \begin{subfigure}[t]{0.49\linewidth}
        \begin{minipage}[b]{1\linewidth}
        \centering
        \includegraphics[width=1\linewidth]{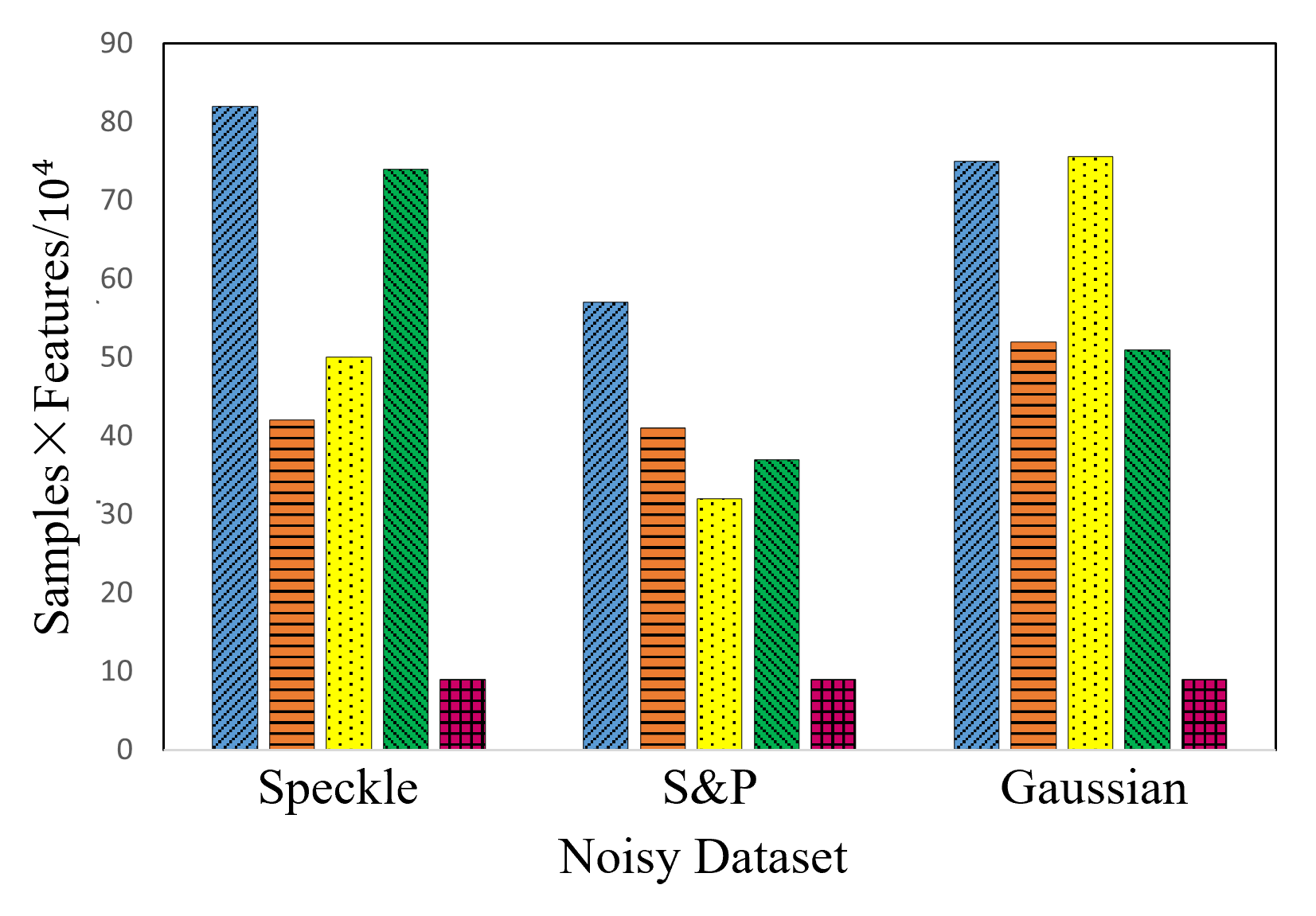}
        \caption{MNIST}
        \end{minipage}
    \end{subfigure}
    \begin{subfigure}[t]{0.49\linewidth}
        \begin{minipage}[b]{1\linewidth}
        \includegraphics[width=1\linewidth]{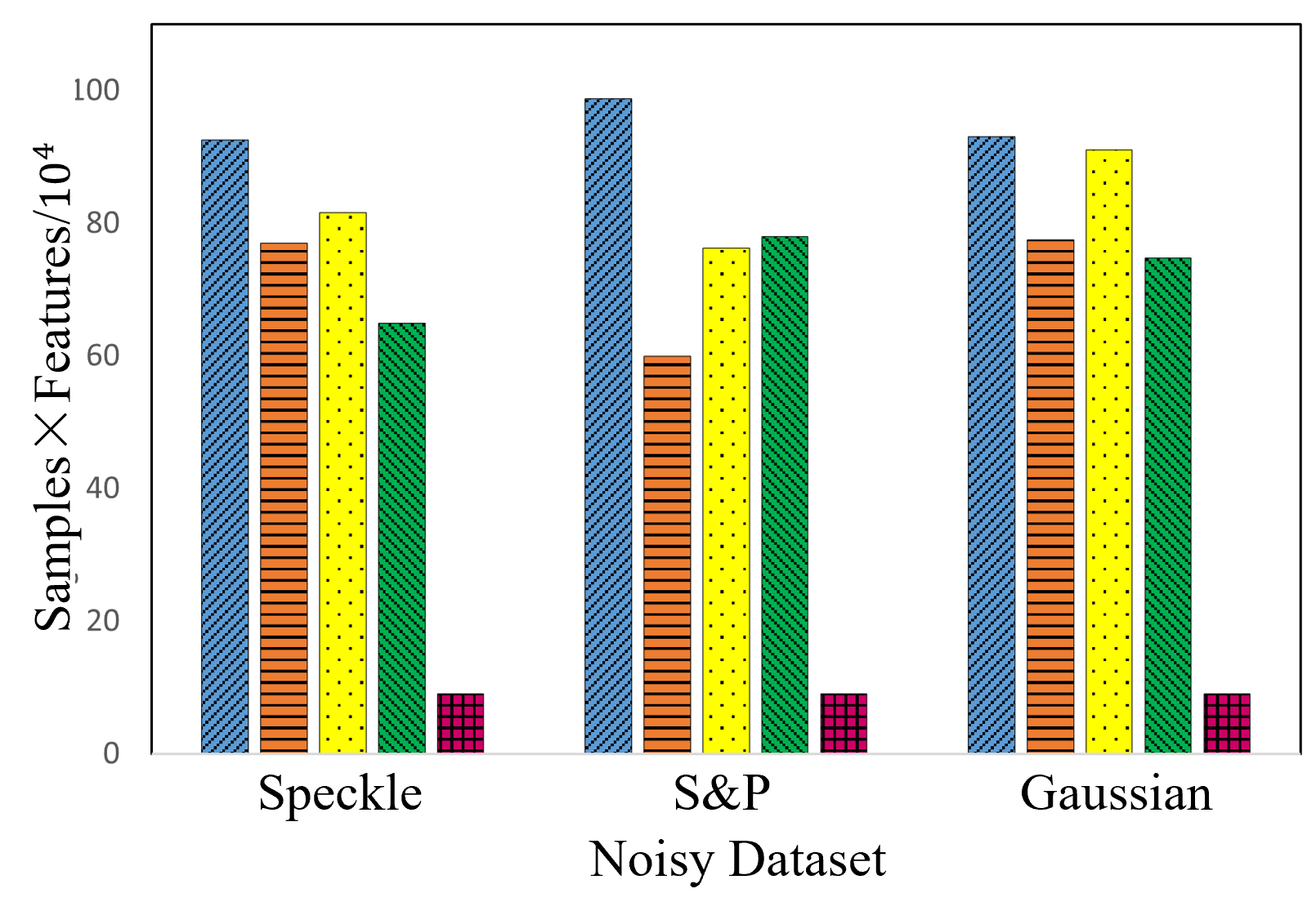}
        \caption{USPS}
        \end{minipage}
    \end{subfigure}
    \begin{subfigure}[t]{0.49\linewidth}
        \begin{minipage}[b]{1\linewidth}
        \includegraphics[width=1\linewidth]{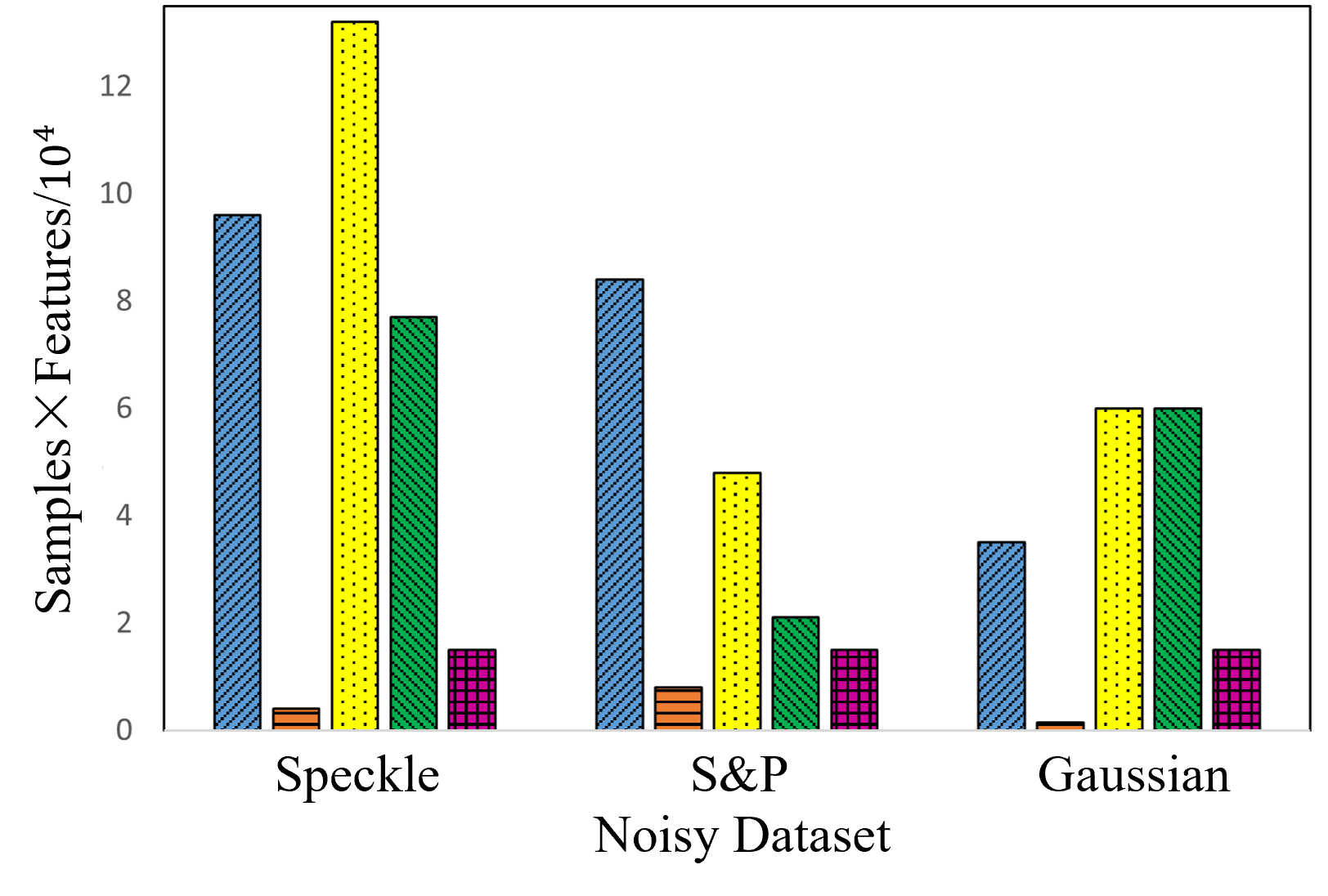}
        \caption{Puish\_url}
        \end{minipage}
    \end{subfigure}
    \begin{subfigure}[t]{0.49\linewidth}
        \begin{minipage}[b]{1\linewidth}
        \includegraphics[width=1\linewidth]{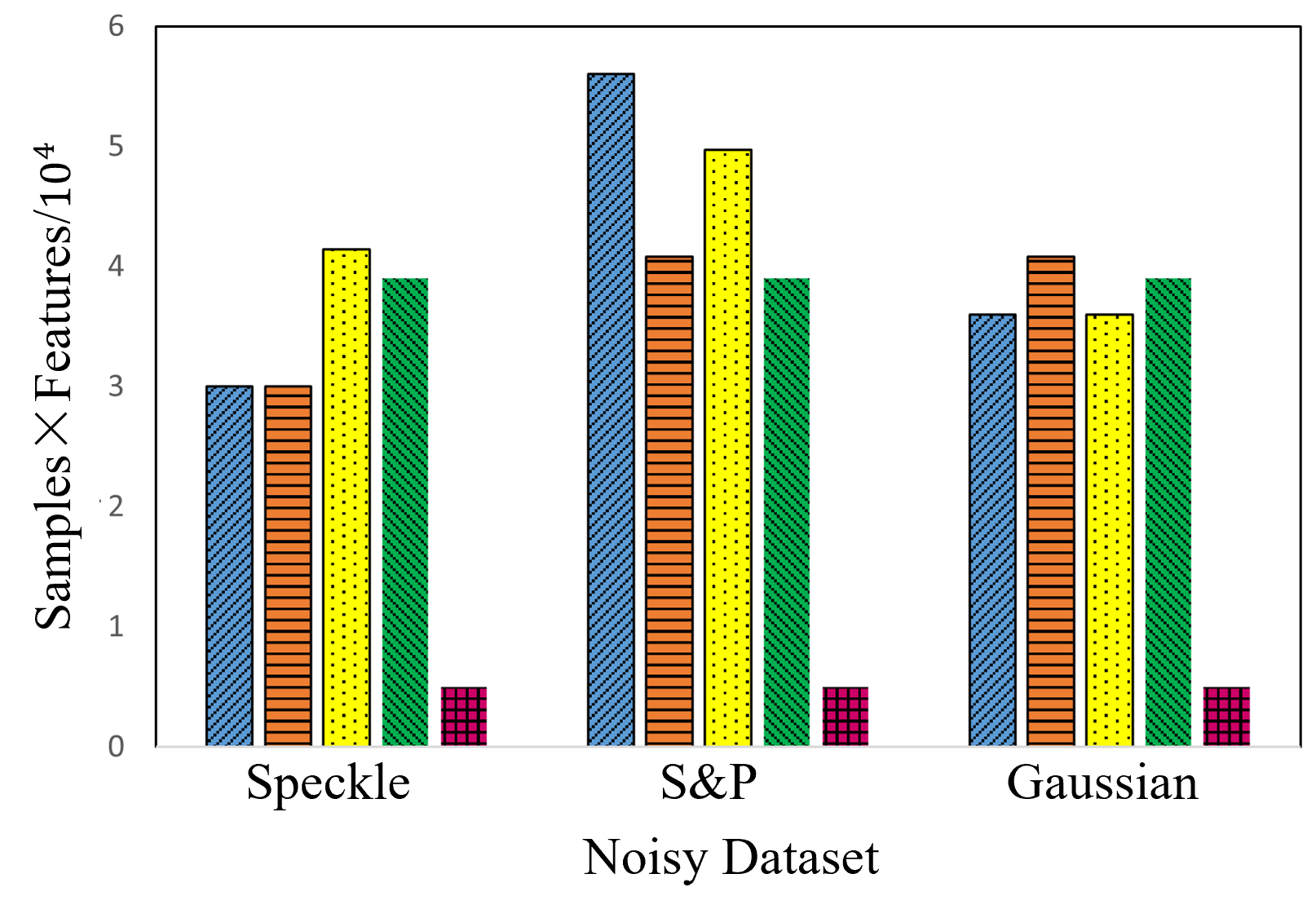}
        \caption{Pendigits}
        \end{minipage}
    \end{subfigure}
    \begin{subfigure}[t]{0.49\linewidth}
        \begin{minipage}[b]{1\linewidth}
        \includegraphics[width=1\linewidth]{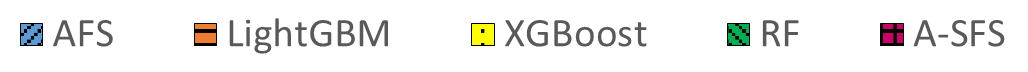}
        \end{minipage}
    \end{subfigure}
    \caption{Compare the number of samples$*$features required for different models to reach the baseline  accuracy}
\label{fig:sampleNums}
\end{figure}

To analyze the model's dependence on label samples in more depth, we compare different supervised feature selection methods for label sample requirements in this section. Based on the accuracy rate that A-SFS can achieve by using 1000/3000 samples to select 5/30 features after the 2000 training step, we compare the number of label samples features required by other methods to achieve this benchmark. Results are shown in Fig. \ref{fig:sampleNums}.

The classification accuracy of different methods might have a very complex pattern regarding the number of samples and TopK features. The comparison result is shown in the histogram below, and more details are shown in the appendix.
We set the features range [30,50] in MNIST and [100,160] in USPS and the number of samples [8000,20000] in MNIST and [5500,7000] in USPS for comparison. The results show that to achieve the same classification effect as A-SFS in the MNIST\_Speckle dataset, AFS needs 20,000 training samples to select the TOP41 Feature number. LGB needs 10,000 training samples to select the TOP42 feature numbers.
A-SFS only needs 10.9\%$\sim$21.4\% of the supervised data of the supervised method required to achieve the benchmark in the MNIST dataset, and only 9.7\%$\sim$13.8\% of the supervised data on the USPS dataset. The reason may be that a large amount of noise data will cause misjudgments and errors in selecting split points for LGB and XGB models. 

For applications that require high-cost investment to obtain high-quality supervised labels in a noisy environment (such as cancer judgment, gene selection, etc.), the potential of self/semi-supervised learning is vast. They can perform dimensionality reduction and importance index evaluation with only a few labeled sample datasets to obtain better downstream performance.
A-SFS is more satisfactory than others in all situations, which benefits from capturing the most critical features carried all features information and the non-linear information integration in the representation.


\subsection{Ablation studies}
\noindent\textbf{Variants performance}

This section design variants and engage ablation experiments to analyze each component's individual or fusion improvement effect in the A-SFS method. 
We define two variants of A-SFS as follows: 

\noindent\textbf{A-SFS$^{-s}$}: Exclude the self-supervised learning part(supervised only).

\noindent\textbf{A-SFS$^{-l}$}: Exclude the self-supervised positioning noise data task(Feature vector estimation self-supervised task only).

\begin{table}[!htb]
 \centering
  \caption{Ablation study in MNIST datasets with the different settings within 3 variants. NM denotes Noise (S\&P)+Missing 30\% data }
    \begin{tabular}{lrrr}
    \toprule
    \textbf{Setting}  & A-SFS &  A-SFS$^{-s}$ & A-SFS$^{-l}$ \\
    \midrule
    Normal         &83.75$\pm$0.25 & \textbf{83.77$\pm$0.23}   & 83.51$\pm$1.87  \\
    Noise(S\&P)    &\textbf{83.62$\pm$0.24} & 81.39$\pm$0.22   & 83.53$\pm$0.18\\
    Missing data   &\textbf{83.76$\pm$0.16} & 81.41$\pm$0.32   & 83.05$\pm$0.25\\
    NM             &\textbf{81.69$\pm$0.96} & 80.94$\pm$1.01   & 80.87$\pm$3.86\\
    \bottomrule
    \end{tabular}
    \label{tab:ablation}
\end{table}%

Results show that self-and Semi-supervised models(both A-SFS and  A-SFS$^{-l}$) achieve performance headways compared with Supervised only(A-SFS$^{-s}$) in noise settings, and the two collaborative act results in the best performance gain than others.
It is worth mentioning that A-SFS$^{-s}$ leads to a more considerable performance drop than A-SFS$^{-l}$. It is because that the supervised-only model(A-SFS$^{-s}$) is trained exclusively on limit labeled samples without the unsupervised information representation and structure mining. In contrast, in the latter, the feature selection model is trained via supervised data and unsupervised representation.

However, under normal settings, the self-supervised mask module has brought missing information to the originally ideal data intentionally, resulting in A-SFS not being as robust as noise settings.
The ablation experiments show that these two self-supervised tasks could enlighten and boost each other in a unified framework. The latent feature structure they learned from different views can help better network structures robust to noisy to guide feature selection.

\noindent\textbf{Variants visualization}

\begin{figure}[h]
\centering
 \includegraphics[width=0.65\columnwidth]{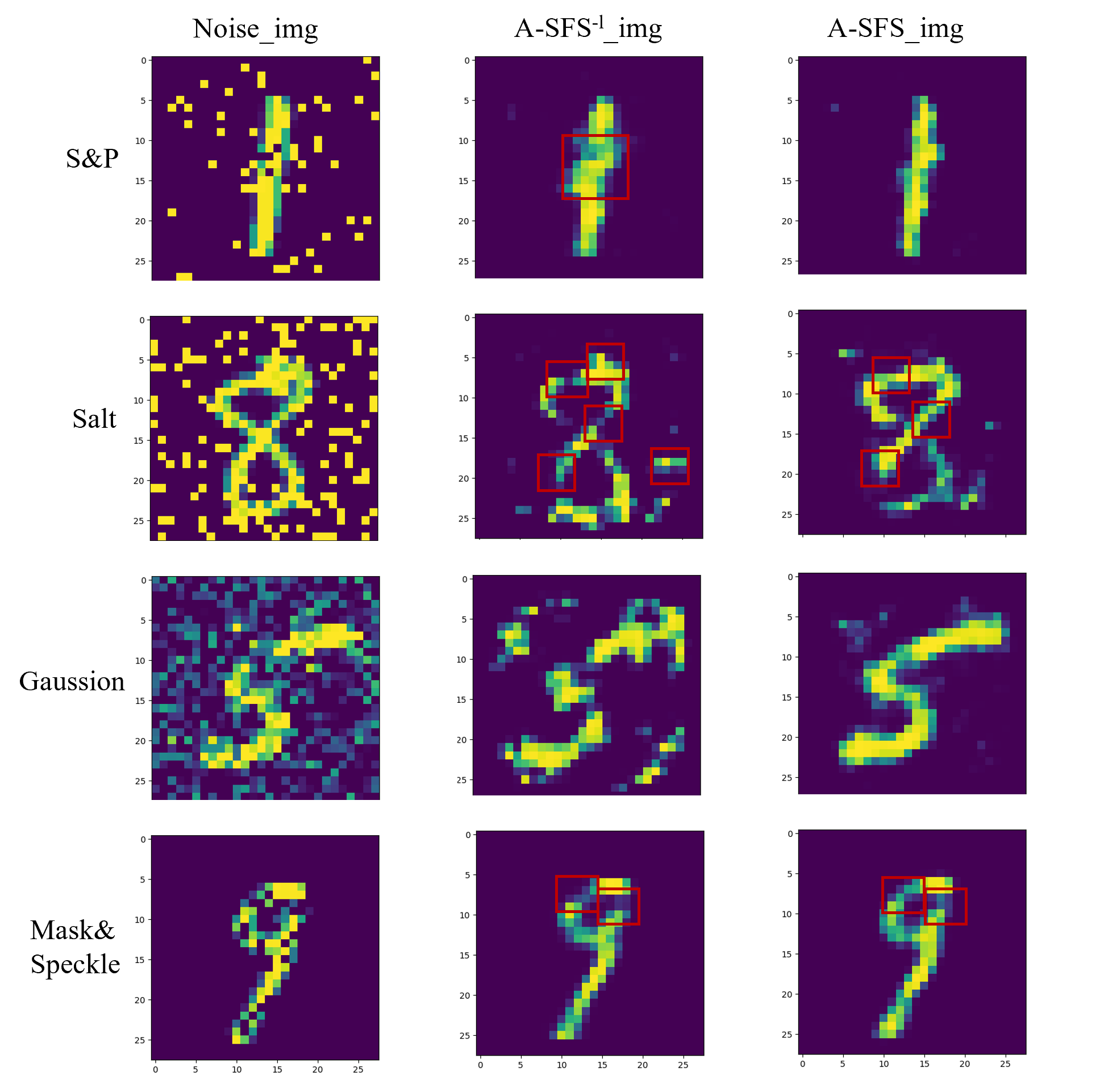}
\centering
\caption{Pre-processed results of A-SFS$^{-l}$ v.s. A-SFS}
\label{fig:Visualization_Ablation}
\end{figure}

In order to further explore the influence of multi-task fusion on structural learning and the guidance of location information on feature learning, we compared the visualization map of A-SFS dual-task self-supervised restoration with the single-task of feature vector estimation (removal of mask vector estimation). 
The result is shown in Fig. \ref{fig:Visualization_Ablation}. The first column is the noise data map. We added S\&P, Salt, Gaussian, Mask\&Speckle noise to the original image. The second and third columns are the recovery graphs of single-task A-SFS$^{-l}$ and dual-task A-SFS. For example, under the S\&P setting, the result of A-SFS$^{-l}$ is far less powerful than that of A-SFS in noise-intensive areas. Under the Salt setting, for some missing pixels, A-SFS can well use the surrounding pixels to recover. As for environmental noise at a long distance, benefiting from the support of location information, it can also better help locate and remove noise data.


Thus, the two tasks can better restore the original data with the joint structure discovery than any of the two single-task learning. Although those self-supervised tasks might have little direct correlation with the modeling target. However, those discovery structural pattern can indeed help the process of feature selection. 




\section{Conclusion}

The non-obviously structured form of tabular data makes deep learning a great success that has not been fully expanded to tabular data. This paper effectively expanding machine learning in complex tabular fields. Based on the combination of self-supervised and attention mechanisms, we propose a new semi-supervised feature selection method. This solution is straightforward and does not need to introduce any assumptions or other objective functions. Extensive experiments prove the effectiveness of this method the accuracy, reduced label dependence, and more robustness towards noise in the data. A promising direction for future work is to design a more powerful model, e.g., a graph neural network with better expressive power for the latent structure estimation.

\section{Acknowledgements}
This work was supported by the National Science Foundation of China (61772473, 62073345 and 62011530148). The computing resources supporting this work were partially provided by High-Flyer AI. (Hangzhou High-Flyer AI Fundamental Research Co., Ltd.)
\bibliographystyle{elsarticle-num} 
\bibliography{asfs}

\section*{Appendix}

\subsection*{Visualization of noise robustness}

To further analyzing the effectiveness of the two pretext tasks, we add various noises: Imputation, Random Gaussian, Salt\&Pepper, Poisson, Speckle, Gaussian Blur and Mean Blur noise to the MNIST datasets, with default noise ratio 0.05. 

\noindent\textbf{Settings of Introduced Noise}
\begin{itemize}
    \item \textbf{Gaussian Noise}: Gaussian-distributed additive noise, default setting: mean=0 and var =0.01;
    \item \textbf{Salt\&Pepper Noise}:High grayscale noise and low grayscale noise appear simultaneously, showing black and white noise. The ratio of high grayscale and low grayscale noise is 0.1;
    \item \textbf{Poisson Noise}: Poisson-distributed noise generated from the data;
    \item \textbf{Speckle Noise}:$out=data+n\times data$ where n is Gaussian noise with specified mean and variance, we set mean=0, var=0.3;
    \item \textbf{Gaussian Blur}:The images convolve with the normal distribution. High-frequency signals are filtered, and low-frequency signals are retained;
    \item \textbf{Mean Blur}:The output pixel value of the filter used in the smoothing process is the mean value of the pixel value in the kernel window (all pixels have the same weighting coefficient).
\end{itemize} 

\begin{figure}
    \centering
    \includegraphics[width=1.\textwidth]{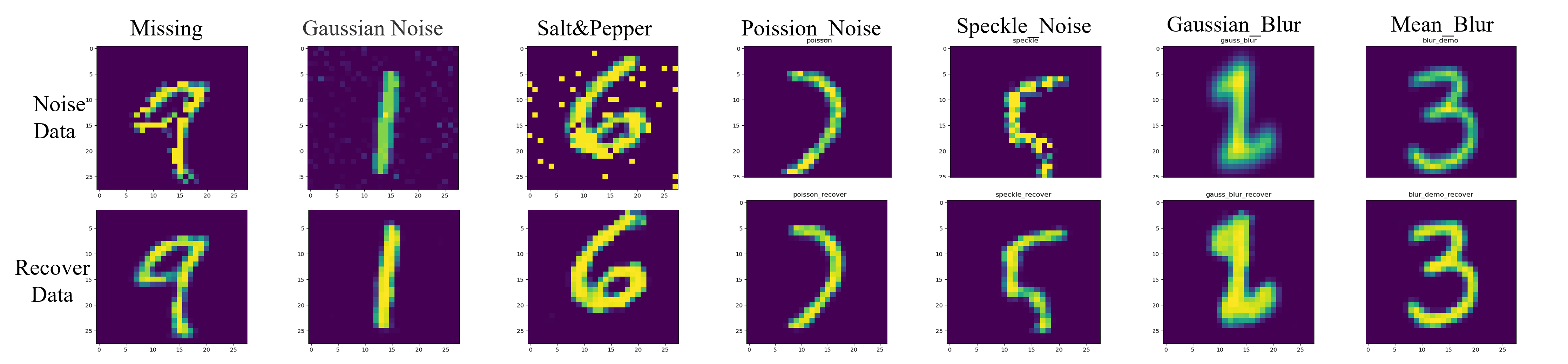}
    \caption{Exemplary data are preprocessing results of the proposed self-supervised module. (Top) The original image of the noise data. (Bottom) The self-supervising module restores the image}
    \label{fig:app_my_visualize}
\end{figure}

It is observed from Figure 1 that in the A-SFS framework structure, self-supervision has robust
characteristics against random noise. It can realize the restoration and restoration of noise or abnormal data. The reason is that the designed self-supervised structure can capture a series of hidden feature knowledge, such as the correlation information, structural relationship, and importance between features. The knowledge of these features helps $S_m$  the decoding process locate the replaced noise data through the non-predictive performance between the feature values. For instance, if the value of a feature is very different from its correlated features, this feature is likely masked and corrupted. And then, $S_r$  calculates the actual value of the replaced noise data through the relevant non-noise features. 

This process is similar to the self-supervised learning in the image and language fields. Images use spatial correlation and location information to determine the angle of rotation, and language uses contextual relevance to synthesize complete word order sentences. Although general tabular data lack explicit spatial correlation and semantic connection, the proposed self-supervised autoencoder can still mines the implicit structural relations between features. Thus, this autoencoder enables the representation of data structures without the needs of capture the information representation of the global structure of the data. It lays the foundation for valuable input for downstream tasks.

\subsection*{Label dependence experiment supplementary data}

we select and compare different supervised feature selection methods for label sample requirements in label dependence experiment. Based on the accuracy rate that A-SFS can achieve by using 1000/3000 samples to select 5/30 features in 3000 training steps, we compare the number of labeled samples and features  required by all methods to achieve this benchmark. The results of the four different datasets are shown in the figure below.

\begin{table}[!htb]
 \centering
  \caption{Comparison of experimental labeled data requirements by different algorithms in MNIST dataset }
    \resizebox{1\columnwidth}{!}{
    \begin{tabular}{lccccccccc}
    \toprule
    \textbf{Noise}  & \multicolumn{3}{|c|}{Speckle} &  \multicolumn{3}{|c|}{S\&P}   &    \multicolumn{3}{|c|}{Gaussian}  \\
    \midrule
    \textbf{Algorithms}  &   samples   &  features & sum$/10^4$  &samples    &  features  &  sum$/10^4$   &samples&      features&sum$/10^4$ \\
    \hline
    AFS         &20000     & 41   &82   & 15000  & 38 &  57   &15000   &50     &   75 \\
    LightGBM	      &10000	&42	     &42	&	10000&	41&	41	&	13000&	40&	52\\
    XGBoost	&10000&	50&	50	&	8000&	40	&32	&	18000&	42&	75.6\\
    RF&	20000&	37&	74	&	10000&	37&	37	&	15000&	34&	51\\
    A-SFS&	3000&	30&	9	&	3000&	30&	9	&	3000&	30&	9\\

    \bottomrule
    \end{tabular}
    \label{tab:ablation_mnist}
    }
\vspace{-0.1in}
\end{table}%

\begin{table}[!htb]
 \centering
  \caption{Comparison of experimental labeled data requirements by different algorithms in USPS dataset }
    \resizebox{1\columnwidth}{!}{
    \begin{tabular}{lccccccccc}
    \toprule
    \textbf{Noise}  & \multicolumn{3}{|c|}{Speckle} &  \multicolumn{3}{|c|}{S\&P}   &    \multicolumn{3}{|c|}{Gaussian}  \\
    \midrule
    \textbf{Algorithms}  &   samples   &  features & sum$/10^4$  &samples    &  features  &  sum$/10^4$   &samples&      features&sum$/10^4$ \\
    \hline
 AFS   & 6300&	147	&92.61	&	6500&	152	&98.8	&	7000&	133	&93.1\\
 LightGBM  & 5500&	140	&77	&	6000&	100&	60&		6800&	114	&77.52\\
 XGBoost  & 6000&	136&	81.6&		6300&	121	&76.23&		7000&	130&	91\\
RF&    5800	&112&	64.96&		6000&	130	&78	&	6500&	115	&74.75\\
      A-SFS&   3000&	30&	9&		3000&	30&	9&		3000&	30&	9\\

    \bottomrule
    \end{tabular}
    \label{tab:ablation_usps}
    }
\vspace{-0.1in}
\end{table}%

\begin{table}[!htb]
 \centering
  \caption{Comparison of experimental labeled data requirements by different algorithms in Puish\_url dataset }
    \resizebox{1\columnwidth}{!}{
    \begin{tabular}{lccccccccc}
    \toprule
    \textbf{Noise}  & \multicolumn{3}{|c|}{Speckle} &  \multicolumn{3}{|c|}{S\&P}   &    \multicolumn{3}{|c|}{Gaussian}  \\
    \midrule
    \textbf{Algorithms}  &   samples   &  features & sum$/10^4$  &samples    &  features  &  sum$/10^4$   &samples&      features&sum$/10^4$ \\
    \hline
 AFS   & 12000&	8  &9.6  &	12000&	7  &8.4    &	7000  &	  5   &   3.5\\
 LightGBM  & 800&	5&	0.4	&2000	&4&	0.8	&500	&3	&0.15\\
 XGBoost  & 12000&	11&	13.2&	8000&	6&	4.8&	10000&	6	&6.0\\
RF&    11000&	7	&7.7&	7000&	3	&2.1&	12000&	5&	6.0\\
      A-SFS&   3000&	5&	1.5&		3000&	5&	1.5&		3000&	5&	1.5\\

    \bottomrule
    \end{tabular}
    \label{tab:ablation_puish}
    }
\vspace{-0.1in}
\end{table}%

\begin{table}[!htb]
 \centering
  \caption{Comparison of experimental labeled data requirements by different algorithms in Pendigits dataset }
    \resizebox{1\columnwidth}{!}{
    \begin{tabular}{lccccccccc}
    \toprule
    \textbf{Noise}  & \multicolumn{3}{|c|}{Speckle} &  \multicolumn{3}{|c|}{S\&P}   &    \multicolumn{3}{|c|}{Gaussian}  \\
    \midrule
    \textbf{Algorithms}  &   samples   &  features & sum$/10^4$  &samples    &  features  &  sum$/10^4$   &samples&      features&sum$/10^4$ \\
    \hline
AFS&	5000&	6	&3&	7000&	8&	5.6&	6000&	6&	3.6\\
lgb&	5000&	6	&3&	6800&	6&	4.08&	6800&	6&	4.08\\
xgb&	6900&	6	&4.14&	7100&	7&	4.97&	6000&	6&	3.6\\
RF&	6500&	6&	3.9&	6500	&6&	3.9&	6500&	6&	3.9\\
A-SFS&	1000&	5&	0.5	&1000&	5&	0.5	&1000&	5&	0.5\\
    \bottomrule
    \end{tabular}
    \label{tab:ablation_pendigits}
    }
\vspace{-0.1in}
\end{table}%

\end{document}